\documentclass[letterpaper, 10 pt, conference]{ieeeconf}  
\IEEEoverridecommandlockouts                              
\usepackage[utf8]{inputenc}
\usepackage[T1]{fontenc}
\usepackage[hidelinks]{hyperref}
\usepackage{graphicx}
\usepackage{tabularx}
\usepackage{amsmath,amssymb}
\usepackage{xspace}
\usepackage[per-mode=symbol,binary-units=true,range-units=single,range-phrase=-,detect-weight=true,detect-family=true]{siunitx}
\usepackage{booktabs}
\usepackage{paralist}
\usepackage{subcaption}
\usepackage{url}
\usepackage{tikz}
\usepackage{pgfplots}
\usetikzlibrary{fit,shapes.callouts,shapes.geometric,backgrounds,positioning,arrows.meta}
\graphicspath{{./figures/}}
\newcommand{\ie}{i.e.,\ }
\newcommand{\eg}{e.g.,\ }
\newcommand{\cf}{cf.\xspace}
\newcommand{\etal}{\xspace{}et al.\xspace}

\newcommand{\reffig}[1]{Fig.~\ref{#1}}
\newcommand{\reftab}[1]{Tab.~\ref{#1}}
\newcommand{\refsec}[1]{Sec.~\ref{#1}}

\DeclareMathOperator{\atantwo}{atan2}
\DeclareMathOperator{\conv}{conv}
\DeclareMathOperator{\relu}{ReLu}
\DeclareMathOperator{\logsoft}{LogSoftmax}
\DeclareMathOperator{\wNLL}{wNLL}
\DeclareMathOperator*{\argmin}{arg\,min}

\title{\LARGE \bf
Visually Guided Balloon Popping with an\\Autonomous MAV at MBZIRC 2020
}

\author{Marius Beul, Simon Bultmann, Andre Rochow, Radu Alexandru Rosu,\\Daniel Schleich, Malte Splietker, and Sven Behnke 
\thanks{This work has been supported by a grant of the Mohamed Bin Zayed International Robotics Challenge (MBZIRC) and the German Federal Ministry of Education and Research (BMBF) in the project ”Kompetenzzentrum: Aufbau des Deutschen Rettungsrobotik-Zentrums (A-DRZ)``}%
\thanks{Institute for Computer Science VI, Autonomous Intelligent Systems, University of Bonn, Endenicher Allee 19a, 53115 Bonn, Germany,
		{\tt\small mbeul@ais.uni-bonn.de}%
}
}

\begin{document}

\maketitle
\thispagestyle{empty}
\pagestyle{empty}

\begin{tikzpicture}[overlay, remember picture]
  \path (current page.north) ++(0.0,-1.0) node[draw = black] {Accepted for IEEE International Symposium on Safety, Security, and Rescue Robotics (SSRR), Abu Dhabi, UAE, 2020};
\end{tikzpicture}
\vspace{-0.3cm}

\begin{abstract}
Visually guided control of micro aerial vehicles (MAV) demands for robust real-time perception, fast trajectory generation, and a capable flight platform.
We present a fully autonomous MAV that is able to pop balloons, relying only on onboard sensing and computing.
The system is evaluated with real robot experiments during the Mohamed Bin Zayed International Robotics Challenge (MBZIRC) 2020 where it showed its resilience and speed. In all three competition runs we were able to pop all five balloons in less than two minutes flight time with a single MAV.
\end{abstract}

\section{Introduction}
\label{sec:Introduction}
In order to advance the state of the art in autonomous mobile robots, the Mohamed Bin Zayed International Robotics Challenge (MBZIRC) 2020 \cite{MBZIRC2020}, which took place in February 2020 in Abu Dhabi posed multiple challenges. One of the tasks was to pop balloons in an outdoor arena of size \SI{90 x 40}{\meter}. Five green balloons with approximately \SI{60}{\centi\meter} diameter were randomly placed inside on top of \SI{2.5}{\meter} long poles. Although the total challenge time was set to \SI{15}{\minute}, the task had to be completed much faster and autonomously to receive a high score. Up to three micro aerial vehicles (MAV) could be used to complete the challenge, but we found it sufficient to use only one MAV. While global navigation satellite system (GNSS) positioning was available, the use of differential GNSS was penalized. In this paper, we present our integrated MAV system ``Jelly'', specifically designed to pop balloons including
\begin{compactitem}
  \item a custom-tailored hardware design,
  \item fast perception accelerated by a Tensor Processing Unit,
  \item robust filtering of sensor data, and
  \item fast trajectory generation and control.
\end{compactitem}

\begin{figure}[t]
  \centering
  \includegraphics[clip,width=1.0\linewidth]{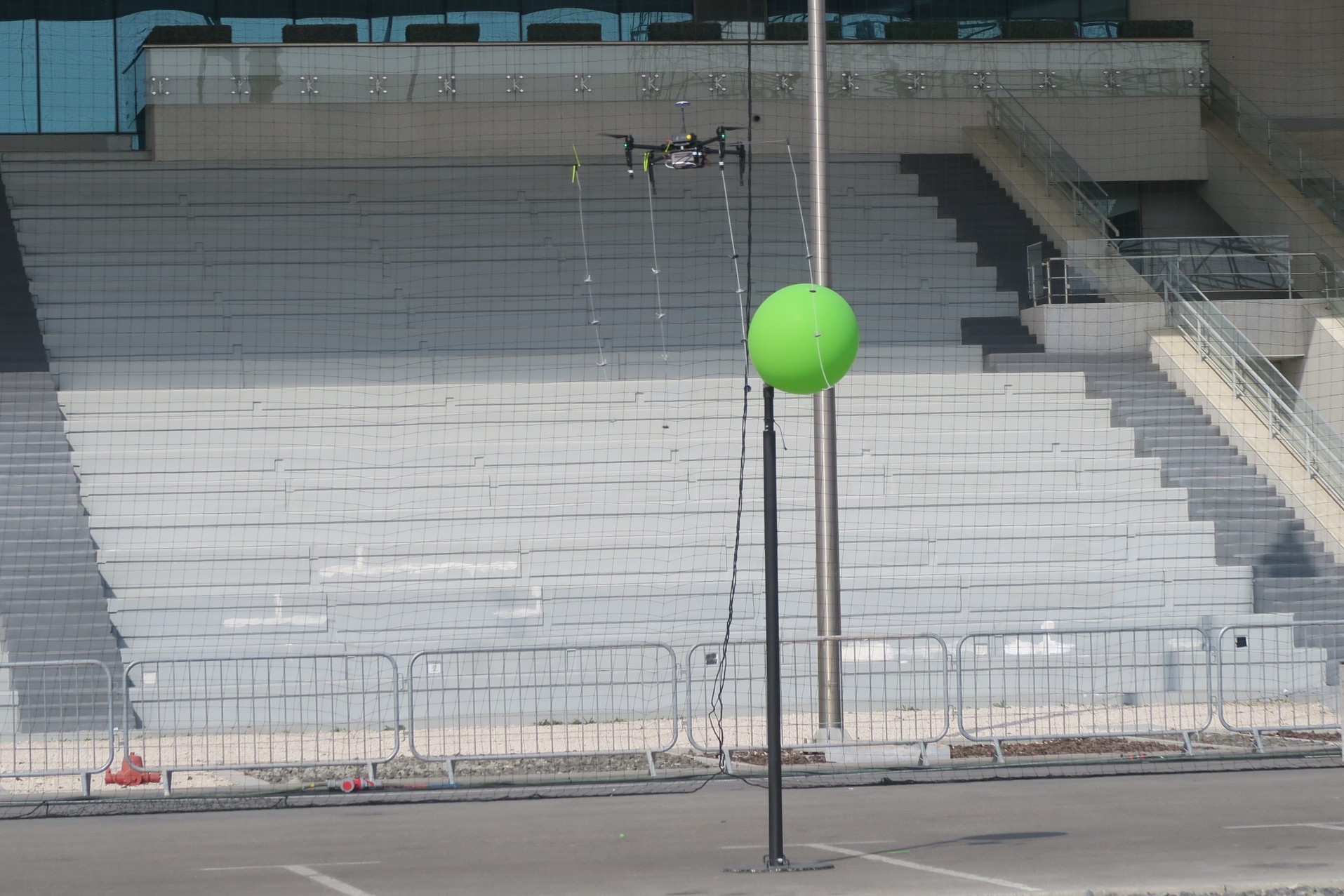}~
  \caption{Our MAV ``Jelly'' just before it pops the balloon with its spiked tentacles.}
  \label{fig:IMG_5629_cut}
  \vspace{-2.5ex}
\end{figure}

We evaluate our approach with real robot experiments and report results from the MBZIRC 2020 competition. \reffig{fig:IMG_5629_cut} shows Jelly popping one balloon with its tentacles.

\section{Related Work}
\label{sec:Related_Work}

At present time, no other group has presented complete systems to this specific task. However, much related research deals with subtasks posed in this MBZIRC 2020 challenge like visual object detection, lightweight computer vision models for deployment on MAVs, or precise MAV control.

Rodriguez et al.~\cite{RoboCup_2019} use a convolutional neural network to not only detect circular objects, but also other pretrained objects in real time.
Circular objects can be easily detected and differentiated from non-circular ones based on the shape of their contour.
Yang et al.~\cite{object_contour_2016} propose an encoder-decoder structure for contour detection of generic foreground objects from the PascalVOC dataset~\cite{pascalvoc_2010}. Maninis et al.~\cite{conv_boundaries_2018} use a CNN architecture to detect object contours at multiple scales together with their orientations, based on a generic backbone CNN, like ResNet~\cite{he_deep_2016}. We follow a similar approach, but train our contour detection network to detect only contours of one class of objects---the balloons (see~\refsec{sec:Balloon_Perception}).

Lightweight computer vision models that can be executed efficiently also on mobile or embedded systems with very restricted computational power have been of increasing research interest during recent years. The MobileNet architectures~\cite{mobilenet_2017,mobilenetv2_2018}, for example, greatly reduce the number of parameters in a convolutional neural network (CNN) by replacing standard convolutions with depthwise-separable convolutions. For our vision system, we employ a standard ResNet architecture but with very few layers (\cf~\refsec{sec:Balloon_Perception}), keeping the number of parameters and the necessary computational power low.
Furthermore, specialized inference accelerators like the Google Edge TPU~\cite{edgetpu_usb} can be used for efficient processing with limited size and energy budget. To make a trained CNN model compatible with the Edge TPU, weights and activations need to be quantized to 8-bit integer values, e.g. using the quantization scheme described in~\cite{quantization_2018}.

Also, fast real-time trajectory generation and control is an active area of research. Specifically, as a result of MBZIRC 2017, various groups presented advanced control approaches for MAVs.
The team from Czech Technical University Prague reports their approaches to landing on a moving platform during the MBZIRC 2017 in~\cite{Baca2017}.
Also, Cantelli \etal and Battiato \etal from the University of Catania report their systems~\cite{Cantelli2017,Battiato2017} including their control approach.
Falanga \etal~\cite{Falanga_SSRR2017} of University of Zurich plan jerk-minimizing trajectories using a fast analytic polynomial generation method similar to ours. Also, outside of MBZIRC, many groups employ polynomial trajectories for MAV control. For a comparison of polynomial trajectory generation algorithms, see the works of Ezair \etal~\cite{Ezair2014}.

\section{System Setup}
\label{sec:System_Setup}
In the following sections, we first describe the hardware of our MAV in \refsec{sec:Hardware}. We continue by presenting the mission control state machine in \refsec{sec:Mission_Control_State_Machine}. Our balloon perception pipeline including an allocentric filter is detailed in \refsec{sec:Balloon_Perception} and \refsec{sec:Balloon_Filter}. GNSS-based state estimation is supported by a laser height filter which we describe in \refsec{sec:Laser_Height_Filter}. Lastly, our trajectory generation and control method is presented in \refsec{sec:Trajectory_Generation_and_Control}.

\subsection{Hardware}
\label{sec:Hardware}

\begin{figure}[t]
  \centering
  \resizebox{1.0\linewidth}{!}{\input{figures/mav.pgf}}
  \caption{Design of our MAV ``Jelly'' equipped with four spiked tentacles, an Intel RealSense D415 camera, a Google Edge TPU, a laser height sensor and a lightweight but powerful onboard computer. Bottom right: closeup of a 3D-printed detachable tentacle clamp with release force set to medium.}
  \label{fig:mav}
  \vspace{-2.5ex}
\end{figure}

Our MAV, shown in \reffig{fig:mav}, is based on the DJI Matrice~100 platform. It is equipped with a small but fast Gigabyte GB-BSi7T-6500 onboard PC with an Intel\textsuperscript{\textregistered} Core\textsuperscript{\texttrademark} i7-6500U CPU running at \SI{2.5/3.1}{\giga\hertz} and \SI{16}{\giga\byte} of RAM.
Balloons are perceived by an Intel RealSense D415 depth camera~\cite{realsense_d415} with the assistance of a Google Edge TPU USB accelerator~\cite{edgetpu_usb}. For precise height estimation, the MAV uses a downwards facing LIDAR-Lite v3~\cite{garmin_lidarlitev3}.

Balloons are punctured with four detachable spiked \SI{1.4}{\meter} long tentacles, mounted on a horizontal bar with a distance of \SI{30}{\centi\meter}. When an adjustable force (set to \SI{\approx 2}{\newton} during the challenge) is applied to a tentacle, \eg by entangling with the poles, it is removed, preventing the MAV from crashing. On each tentacle, four needle-spiked hemispheres are mounted with \SI{15}{\centi\meter} distance. By using flexible popping hardware, our MAV complied to the size restrictions of \SI{1.2 x 1.2 x 0.5}{\meter}, still offering a forgiving popping system that does not require centimeter-level precision.

\begin{figure}[t]
  \centering
  \resizebox{1.0\linewidth}{!}{\begin{tikzpicture}[font=\sffamily,>={Stealth[inset=0pt,length=4pt,angle'=45]}]
\tikzset{content_node/.append style={minimum size=1.5em,minimum width=6em,draw,align=center,rounded corners,scale=0.65}}
\tikzset{label_node/.append style={scale=0.5}}
\tikzset{group_node/.append style={dotted,align=center,rounded corners,inner sep=1em,thick}}

\definecolor{red}{rgb}     {0.5,0.0,0.0}
\definecolor{green}{rgb}   {0.0,0.5,0.0}
\definecolor{blue}{rgb}    {0.0,0.0,0.5}
\definecolor{grey}{rgb}    {0.5,0.5,0.5}

\draw[thick, rounded corners, grey!20!white,fill] (-4.0,0.5) -- (4,0.5) -- (4.0,4.0) -- (-4.0,4.0) -- cycle;
\draw[thick, rounded corners, grey!20!white,fill] (-4.0,-1.25) -- (4,-1.25) -- (4.0,-0.25) -- (-4.0,-0.25) -- cycle;

\node(Camera)[content_node,fill=green!15!white] at (-3.0,3.5) {Camera};
\node(Balloon_Detection)[content_node,fill=blue!15!white] at (0.0,3.5) {Balloon Detection};
\node(Filter)[content_node,fill=blue!15!white] at (3.0,3.5) {Filter};
\node(State_Machine)[content_node,fill=blue!15!white] at (0.0,2.25) {State Machine};
\node(Trajectory_Generation)[content_node,fill=blue!15!white] at (0.0,1.0) {Trajectory Generation};
\node(Operator)[content_node,fill=green!15!white] at (-3.0,2.875) {Operator};

\node(MAV)[content_node,fill=red!15!white] at (0.0,-1.25+0.5) {MAV};
\node(GNSS)[content_node,fill=green!15!white] at (-3.0,-1.0+0.5) {GNSS};
\node(IMU)[content_node,fill=green!15!white] at (-3.0,-1.5+0.5) {IMU};

\draw[->, thick] (Camera) -- node[label_node,midway,below] {\SI{30}{\hertz}} node[label_node,midway,above] {Image} (Balloon_Detection);
\draw[->, thick] (Balloon_Detection) -- node[label_node,midway,below] {\SI{20}{\hertz}} node[label_node,midway,above] {3D~Position} (Filter);
\draw[->, thick] (Filter) -- (Operator -| Filter) -- (Operator -| State_Machine.30) node[label_node,midway,above,align=center] {3D~filtered~Position} node[label_node,midway,below,align=center] {\SI{20}{\hertz}} --  (State_Machine.30);
\draw[->, thick] (State_Machine) -- node[label_node,midway,left] {\SI{40}{\hertz}} node[label_node,midway,right,align=left] {3D~Target~Position\\Target~Yaw} (Trajectory_Generation);
\draw[->, thick] (Trajectory_Generation) -- node[label_node,midway,left] {\SI{50}{\hertz}} node[label_node,midway,right,text width=1cm] {Roll Pitch Climb~rate Yaw~rate} (MAV);
\draw[->, thick] (MAV) -- (2.25,-1.25+0.5) -- node[label_node,midway,left] {\SI{50}{\hertz}} node[label_node,midway,right,text width=2.8cm] {3D~MAV~Position 3D~MAV~Velocity 3D~MAV~Accel. MAV~Yaw} (2.25,1.0) -- (Trajectory_Generation);
\draw[->,thick] (2.25,1.0) -- (State_Machine -| 2.25,1.0) node[label_node,midway,left] {\SI{50}{\hertz}} node[label_node,midway,right,align=left] {3D~MAV~Position\\MAV~Yaw} -- (State_Machine);

\draw[->, thick] (GNSS) -- (GNSS -| -1.5,-1.0)  -- (-1.5,-1.0 |- MAV.175) -- node[label_node,midway,left] {} node[label_node,midway,right] {} (MAV.175);
\draw[->, thick] (IMU) --  (IMU -| -1.5,-1.0) --  (-1.5,-1.0 |- MAV.185)  -- node[label_node,midway,left] {} node[label_node,midway,right] {} (MAV.185);
\draw[->, thick] (Operator) --  node[label_node,midway,above] {Start/Stop Command} (Operator -| State_Machine.150) -- node[label_node,midway,left] {} node[label_node,midway,right] {} (State_Machine.150);


\node(ROS_Group_Label)[label_node,anchor=south west] at (-4.0,4.0) {\textbf{Onboard Computer}};
\node(MAV_Group_Label)[label_node,anchor=south west] at (-4.0,-0.25) {\textbf{DJI Matrice 100}};

\end{tikzpicture}}
  \caption{Structure of our method. Green boxes represent external inputs like sensors, blue boxes represent software modules, and the red box indicates the MAV flight control. All software components use ROS as middleware. Position, velocity, acceleration, and yaw are allocentric.}
  \label{fig:structure}
  \vspace{-2.5ex}
\end{figure}

For allocentric localization and state estimation, we employ the filter onboard the DJI flight control that incorporates GNSS and IMU data.

To make all components easily transferable between the test area at our lab and also different arenas on site, we defined all coordinates (x, y, z, yaw) in a field-centric coordinate system. The center and orientation of the current field were broadcasted by a base station PC to the MAV. Since we do not make any assumption about the allocentric movement of the target, navigation is purely relative to the detected target and not affected by global inaccuracies.
In contrast to other teams, we did not use advanced satellite-based localization methods like Real Time Kinematic positioning (RTK-GPS) that need multiple GPS antennas on the MAV.

Fig.~\ref{fig:structure} gives an overview of the information flow in our system.
We use the robot operating system (ROS) as middleware on the MAV and the ground control station. We communicate over WiFi with a robust UDP protocol, developed for connections with low bandwidth and high latency~\cite{nimbro_networking}.

\subsection{Mission Control State Machine}
\label{sec:Mission_Control_State_Machine}
\begin{figure}[t]
  \centering
  \resizebox{1.0\linewidth}{!}{\begin{tikzpicture}[font=\sffamily,on grid,>={Stealth[inset=0pt,length=4pt,angle'=45]}]
\tikzset{every node/.append style={node distance=3.0cm}}
\tikzset{terminal_node/.append style={minimum size=1.5em,minimum height=3em,minimum width={width("Search Point")+0.2em},draw,align=center,rounded corners,scale=0.65}}
\tikzset{content_node/.append style={minimum size=1.5em,minimum height=3em,minimum width={width("Search Point")+0.2em},draw,align=center,scale=0.65,fill=blue!15!white}}
\tikzset{label_node/.append style={scale=0.5, near start}}
\tikzset{group_node/.append style={align=center,rounded corners,inner sep=1em,thick}}
\tikzset{decision_node/.append style={align=center,scale=0.5,shape aspect=1.5,minimum width=7.9em,minimum height=5.4em,diamond,draw,fill=yellow!25!white,font=\sffamily\normalsize,node distance=3.9cm}}

\definecolor{red}{rgb}     {0.5,0.0,0.0}
\definecolor{green}{rgb}   {0.0,0.5,0.0}
\definecolor{blue}{rgb}    {0.0,0.0,0.5}
\definecolor{grey}{rgb}    {0.5,0.5,0.5}

\draw[thick, rounded corners, grey!20!white,fill] (1, -2.1) -- (4.75, -2.1) -- (4.75, -0.65) -- (7.0, -0.65) -- (7.0, 0.9) -- (1.0, 0.9) -- cycle;

\draw[thick, rounded corners, grey!20!white,fill] (4.95, -2.1) -- (11.1, -2.1) -- (11.1, 0.9) -- (7.2, 0.9) -- (7.2, -0.85) -- (4.95, -0.85) -- cycle;

\node(takeoff)[terminal_node,fill=red!15!white] at (0, 0) {Takeoff};
\node(fly_to_waypoint)[content_node, right of=takeoff] {Fly to\\Waypoint};
\node(waypoint_reached)[decision_node, right of=fly_to_waypoint] {Waypoint\\Reached?};
\node(set_next_waypoint)[content_node, below of=waypoint_reached, node distance=2.0cm] {Set Next\\Waypoint};
\node(has_balloon_detection)[decision_node, right of=waypoint_reached] {Has Balloon\\Detection?};
\node(pop_balloon)[content_node, right of=has_balloon_detection, node distance=3.7cm] {Pop Balloon};
\node(target_lost)[decision_node, below of=pop_balloon, node distance=2.6cm] {Target\\Lost?};
\node(balloon_popped)[decision_node, right of=target_lost] {Balloon\\Popped?};
\node(return_to_center)[content_node, left of=target_lost, node distance=3.7cm, densely dashed] {Return to\\Center};

\draw[->, thick] (takeoff) -- (fly_to_waypoint);
\draw[->, thick] (fly_to_waypoint) -- (waypoint_reached);
\draw[->, thick] (waypoint_reached) -- node[label_node,right] {Yes} (set_next_waypoint);
\draw[->, thick] (waypoint_reached) -- node[label_node,above] {No} (has_balloon_detection);
\draw[->, thick] (set_next_waypoint) -| (fly_to_waypoint);
\draw[->, thick] (has_balloon_detection.north) -- node[label_node,right, midway] {No} ++(0, 0.2) -| (fly_to_waypoint);
\draw[->, thick] (has_balloon_detection.east) -- node[label_node,above] {Yes} ++(0.3, 0) -- (pop_balloon);
\draw[->, thick] (pop_balloon) -- (target_lost);
\draw[->, thick] (target_lost.west) -- node[label_node,above] {Yes} ++(-0.3, 0) -- (return_to_center);
\draw[->, thick] (target_lost) -- node[label_node,above] {No} (balloon_popped);
\draw[->, thick] (balloon_popped.north) -- node[label_node,right] {No} ++(0, 0.3) |- (pop_balloon);
\draw[->, thick] (balloon_popped.south) -- node[label_node,right, midway] {Yes} ++(0, -0.2) -| (return_to_center);
\draw[->, thick] (return_to_center) -- (has_balloon_detection);

\node[scale=0.65, anchor=north west] at (1, 0.9) {\textbf{Search}};
\node[scale=0.65, anchor=north west] at (7.2, 0.9) {\textbf{Pop}};

\end{tikzpicture}}
  \caption{The flowchart of our state machine.}
  \label{fig:state_machine}
  \vspace{-2.5ex}
\end{figure}
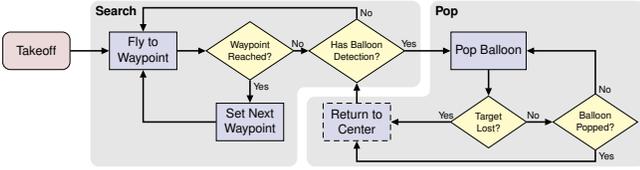

The behavior of the MAV is controlled by a state machine that serves as a generator for waypoints and headings for the subsequent control layers. It also ensures that the MAV does not exceed arena limits and stays within a defined altitude corridor, so that it stays always above the balloon mounting poles and below the \SI{5.0}{\meter} minimum altitude of the other subchallenge's MAVs. \reffig{fig:state_machine} shows a flowchart of our state machine, which consists of two alternating parts---Search and Pop. In search mode, the MAV flies a repeating creeping-line pattern along the long axis of the field, thereby scanning the entire arena. In Pop mode, the MAV flies a trajectory that drags the tentacles through detected balloons.

The MAV velocity in search mode is tuned to \SI{5.0}{\meter\per\second} and the altitude is \SI{4.0}{\meter}, so that balloons can be reliably searched from a safe height. Our balloon detector produces reliable position estimates at ranges over \SI{30}{\meter}. During the Grand Challenge, the search pattern comprised two search lanes, spaced at \SI{10}{\meter} from the arena limits, which proved sufficient. As shown in \reffig{fig:structure}, all balloon detections are filtered (\refsec{sec:Balloon_Filter}) before being processed by the state machine. The filter provides a list of verified balloon positions which are within the arena limits, including those which are currently out of view. Once the state machine receives at least one detection, it proceeds to approach the closest target.

In Pop mode, a straight-line trajectory is computed, such that the center of the tentacles passes through the balloon center at a non-zero velocity. The tentacles' upwards facing needles are dragged into the balloon surface, effectively puncturing it. As the forward-facing camera cannot perceive the balloon all the way, it is assumed to be popped once the MAV passes over the estimated center of the balloon instance within \SI{0.5}{\meter} radius. Should the balloon be still intact, due to unsuccessful puncturing or missing the intercept point, it will be tackled again later as the search pattern repeats and the balloon will inevitably be re-detected. If the target is lost during the approach, e.g. because the filter discarded a false positive, the attempt is cancelled. As an addition in the Grand Challenge, after each attempt to pop a balloon, the MAV returns to the center of the field in order to prevent flying into the scaffolding protruding into the arena. This method of handling the non-convexity of the field is simple but it introduces additional flying time as compared to real obstacle avoidance. On the other hand, it is easy to implement and reliable. After returning to the center, the MAV targets the subsequent closest balloon provided by the filter or resumes with the search pattern, if there are no viable balloon hypotheses.

\subsection{Balloon Perception}
\label{sec:Balloon_Perception}

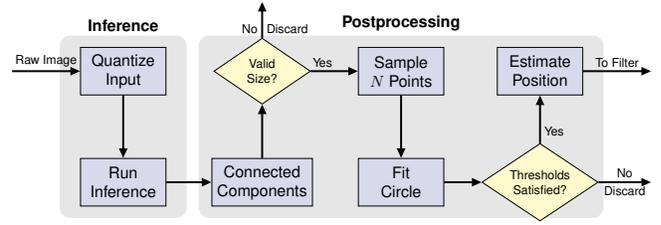
\begin{figure}[t]
  \centering
  \resizebox{1.0\linewidth}{!}{\begin{tikzpicture}[font=\sffamily,on grid,>={Stealth[inset=0pt,length=4pt,angle'=45]}]
\tikzset{every node/.append style={node distance=3.0cm}}
\tikzset{terminal_node/.append style={minimum size=1.5em,minimum height=3em,minimum width={width("Search Point")+0.2em},draw,align=center,rounded corners,scale=0.65}}
\tikzset{content_node/.append style={minimum size=1.5em,minimum height=3em,minimum width={width("Search Point")+0.2em},draw,align=center,scale=0.65,fill=blue!15!white}}
\tikzset{label_node/.append style={scale=0.5, near start}}
\tikzset{group_node/.append style={align=center,rounded corners,inner sep=1em,thick}}
\tikzset{decision_node/.append style={align=center,scale=0.5,shape aspect=1.5,minimum width=7.9em,minimum height=5.4em,diamond,draw,fill=yellow!25!white,font=\sffamily\normalsize,node distance=3.9cm}}

\definecolor{red}{rgb}     {0.5,0.0,0.0}
\definecolor{green}{rgb}   {0.0,0.5,0.0}
\definecolor{blue}{rgb}    {0.0,0.0,0.5}
\definecolor{grey}{rgb}    {0.5,0.5,0.5}

\draw[thick, rounded corners, grey!20!white,fill] (-3.9, 1.3) -- (-2.1, 1.3) -- (-2.1,-1.3) -- (-3.9,-1.3) -- cycle;
\draw[thick, rounded corners, grey!20!white,fill] (-1.9, 1.3) -- ( 3.9, 1.3) -- ( 3.9,-1.3) -- (-1.9,-1.3)  -- cycle;

\node(quantize_input)[content_node] at       (-3.0, 0.8) {Quantize\\Input};
\node(run_inference)[content_node]  at       (-3.0,-0.8) {Run\\Inference};
\node(connected_components)[content_node] at (-1.0,-0.8) {Connected\\Components};
\node(valid_size)[decision_node] at          (-1.0, 0.8) {Valid\\Size?};
\node(sample_x_points)[content_node] at      ( 1.0, 0.8) {Sample\\$N$ Points};
\node(fit_circle)[content_node] at           ( 1.0,-0.8) {Fit\\Circle};
\node(thresholds)[decision_node] at     ( 3.0,-0.8) {Thresholds\\Satisfied?};
\node(estimate_position)[content_node] at    ( 3.0, 0.8) {Estimate\\Position};

\draw[->, thick] (-4.6, 0.8) -- node[label_node,midway,above] {Raw Image} (quantize_input);
\draw[->, thick] (quantize_input) -- (run_inference);
\draw[->, thick] (run_inference) -- (connected_components);
\draw[->, thick] (connected_components) -- (valid_size);
\draw[->, thick] (valid_size) -- node[label_node,above] {Yes} (sample_x_points);
\draw[->, thick] (valid_size) -- node[label_node,left] {No} node[label_node,right] {Discard} (-1.0, 1.8);
\draw[->, thick] (sample_x_points) -- (fit_circle);
\draw[->, thick] (fit_circle) -- (thresholds);
\draw[->, thick] (thresholds) -- node[label_node,right] {Yes} (estimate_position);
\draw[->, thick] (thresholds) -- node[label_node,midway,above] {No} node[label_node,midway,below] {Discard} ( 4.6,-0.8);
\draw[->, thick] (estimate_position) -- node[label_node,midway,above] {To Filter} ( 4.6, 0.8);

\node[scale=0.65, anchor=south] at (-3.0, 1.3) {\textbf{Inference}};
\node[scale=0.65, anchor=south] at ( 1.0, 1.3) {\textbf{Postprocessing}};

\end{tikzpicture}}
  \caption{Perception pipeline: Input is the raw image and output is a number of balloon detections with 3D-positions in the camera coordinate system.}
  \label{fig:det_pipe}
  \vspace{-2.5ex}
\end{figure}

\begin{figure}[t]
    \centering
    \begin{subfigure}[t]{0.49\linewidth}
        \centering
        \includegraphics[width=\linewidth]{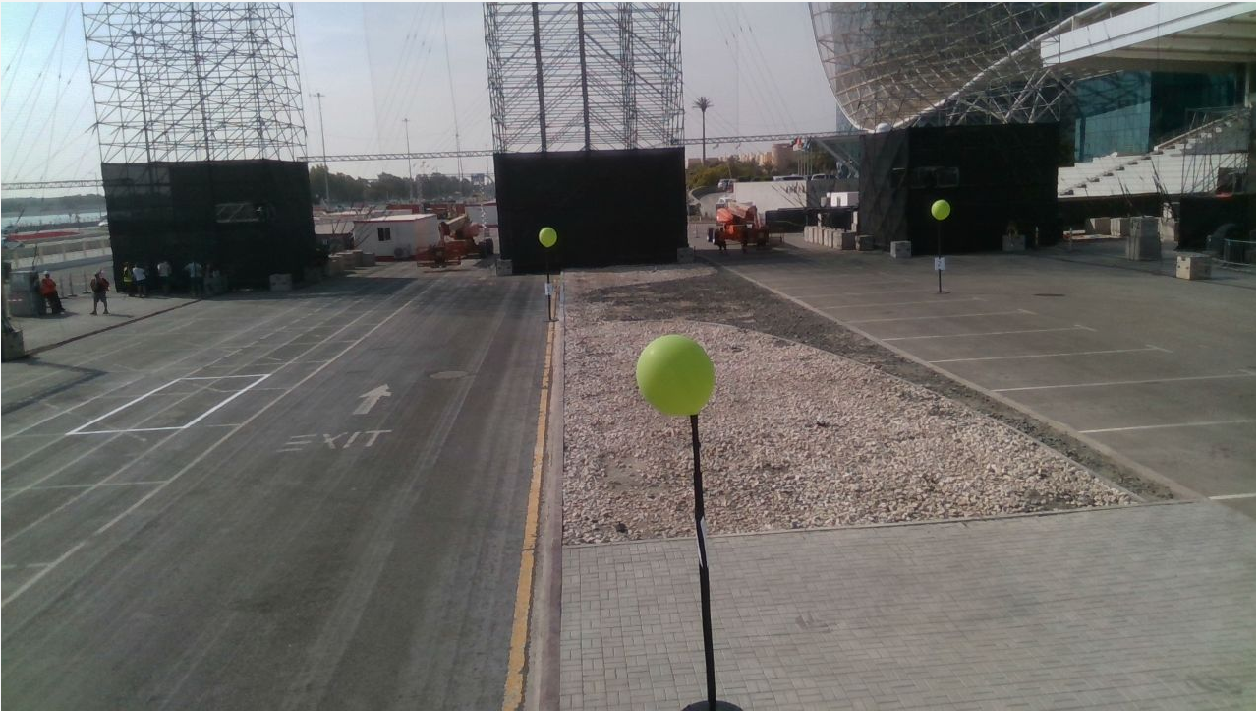}
        \caption{Input image}
    \end{subfigure}
    \begin{subfigure}[t]{0.49\linewidth}
        \centering
        \includegraphics[width=\linewidth]{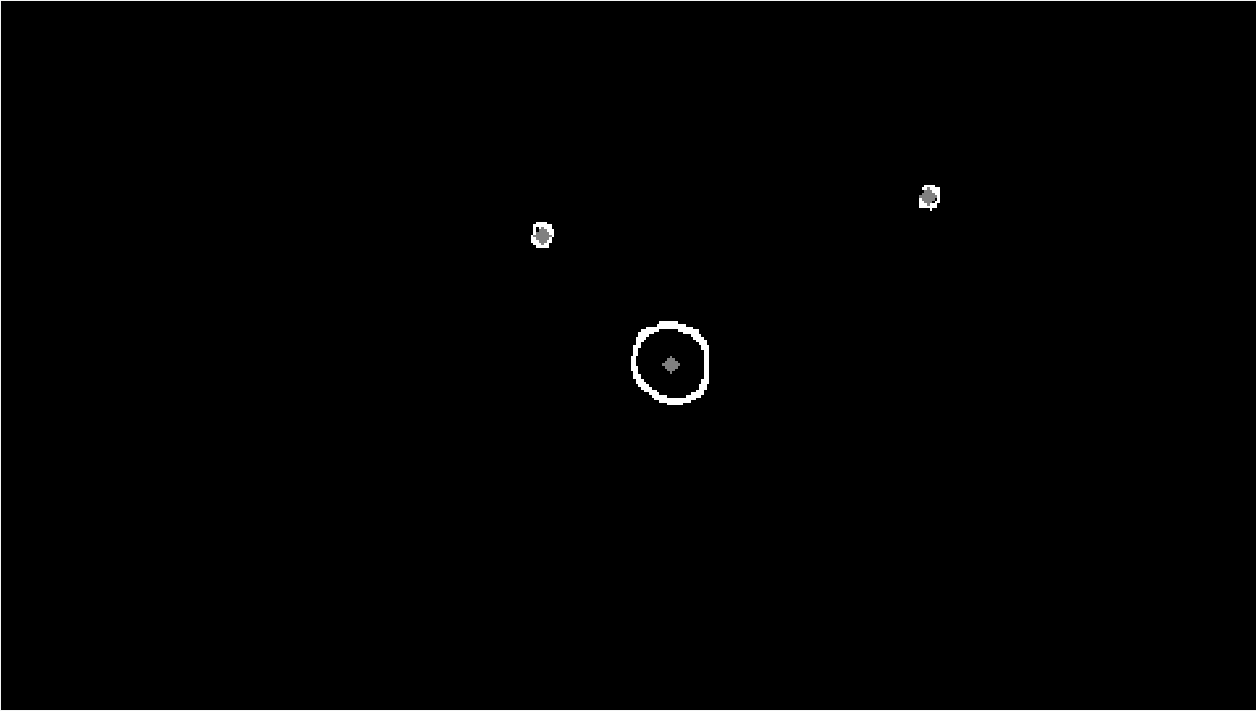}
        \caption{Balloon detections}
    \end{subfigure}
    \caption{Balloon perception: The detected balloon outlines are drawn with white lines and the centers are marked with gray points. All visible balloons are detected, even at large distances, without false positives in the background.}
    \label{fig:balloon_perception}
    \vspace{-2.5ex}
\end{figure}

Our approach for detecting balloons in images is based on deep learning methods and split into an inference and a postprocessing step (\cf~\reffig{fig:det_pipe}). During the inference step a neural network for semantic segmentation is employed. Since we aim to detect multiple balloons, we perform a binary segmentation of the raw input image and extract the balloon outlines as shown in \reffig{fig:balloon_perception}. The balloon detection itself is carried out in the postprocessing step, which provides information like the number of balloons as well as their confidence values and positions in camera coordinates.

\begin{figure}[t]
  \centering
  \includegraphics[trim=00mm 00mm 00mm 00mm,clip,width=1.0\linewidth]{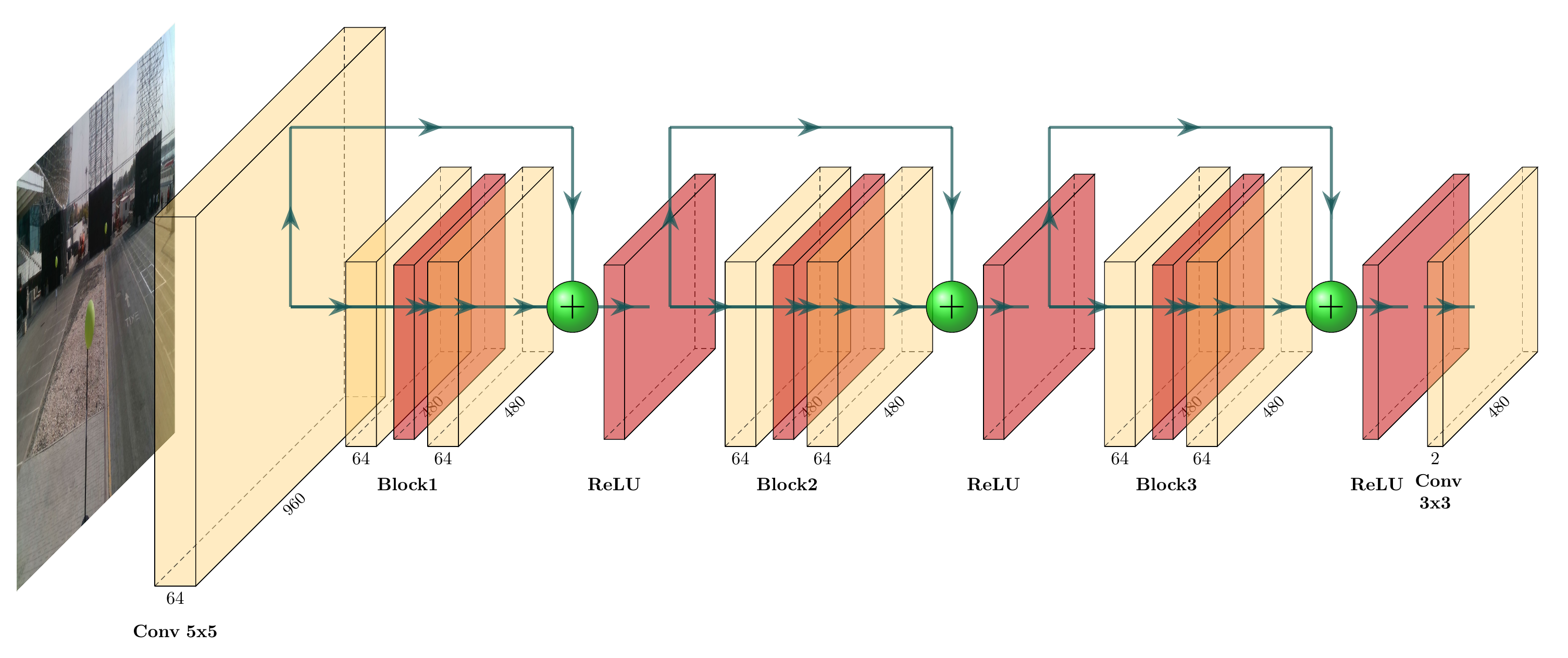}
  \vspace*{-3ex}
  \caption{Architecture of the ball outline segmentation network.}
  \label{fig:NN}
  \vspace{-2.5ex}
\end{figure}

\paragraph{Balloon Outline Segmentation Network}
The structure of the neural network is simple and based on the first three blocks of ResNet-18~\cite{he_deep_2016}. See \reffig{fig:NN} for a visualization of our network. Each block consists of two 3$\times$3 convolutions and calculates 
\begin{align}
x_{\text{out}, i} = \relu\left(x_i+\conv_2\left(\relu\left(\conv_1\left(x_i\right)\right)\right)\right)\,.
\end{align}
The last convolutional layer performs a reduction from 64 to $C=2$ feature maps $f_c, c \in \left\{0,\ldots,C-1\right\}$. These feature maps are used for a binary classification consisting of background and balloon outlines classes. The network is trained with weighted Negative Log-Likelihood Loss ($\wNLL$), after calculating the pixel-wise $\logsoft$ from the output feature maps $f_c$:\vspace*{-2ex}
\begin{align}
q_{i,c} = \logsoft\left(x_{i,c}\right) &= \log\left( \frac{e^{x_{i,c}}}{\sum\limits_{c'=0}^{C-1} e^{x_{i,c'}}}\right)\,,\\
\wNLL(x_i,y_i) &= -w_{y_i}  q_{i,y_i}\,,
\end{align}
for every pixel $x_{i,c}$ in the feature maps $f_c$ and ground truth label $y_i$. The weights $w_c$ are used to compensate the class imbalance of the training set. 

The weights of the first convolutional layer are initialized from ImageNet-pretrained ResNet-18. The network uses a receptive field of 5 and a stride of 2 to sample down the input image, since ResNet uses time consuming convolutions and inference has to run in real time. This initialization results in faster training and better inference results than random initialization.
For training, a dataset consisting of 10,000 synthetic and 300 real images was used, mostly consisting of single balloons and just a few of them containing multiple balloons. The synthetic images were generated by a lightweight physically-based renderer~\cite{easy_pbr}. A 3D mesh of a \SI{60}{\centi\meter} diameter sphere is randomly placed in different HDR environments in order to capture realistic lighting as shown in \reffig{fig:ballon_synthetic}. To reduce the number of false positives, the images also include spherical shapes that are not green and do not correspond to a balloon. The real images were recorded with the same Intel RealSense D415 camera which was used during the competition. To enhance generalization of the network even further, we added noise to the synthetic data as described in~\cite{Carlson:ECCV18}.

The network was trained with image size 960$\times$540. The ground truth size is 480$\times$270 and shows white balloon outlines with a thickness of 1 pixel on a black background. However, we decided to use a 3$\times$3 dilation on the output image to enhance chances to detect a balloon as a fully connected component and make the detections more invariant to noise. Training was accomplished with a batch size of $B=12$ and a learning rate of $lr = 0.001$ using the Adam optimizer on a Nvidia GeForce GTX 1080 Ti. In total, pretraining took 130 epochs, followed by finetuning during the competition for another 50 epochs. On site, we added 250 additional annotated images, which our MAV captured during rehearsals, resulting in a total training time of \SI{9.5}{\hour}.

\paragraph{Postprocessing}
The purpose of the postprocessing step is to detect balloons in the binary segmentation output as shown in \reffig{fig:balloon_perception}. In the best case, the binary segmentation provides all outlines of the balloons with no noise in the background.
Since the circle Hough transform is very time consuming and inefficient for oval shapes (like balloons often are) we use a pipeline which processes the segmentation output in several steps.
In the first step, connected components are extracted and valid components are filtered by a minimum number of pixels. Furthermore, $N$ points are sampled equally distributed on each connected component, starting with the largest one descending. To fit a circle into the sampled points $x_i, i\in\left\{1,\ldots,N\right\}$ we estimate a center point $c$ and the radius
\begin{align}
R_{\text{mean}}(c) = \frac{1}{N}\sum\limits_{i=1}^N{\|x_i-c\|}\,.
\end{align}
To estimate the optimum circle center $\hat{c}$ we minimize the residuals\vspace*{-2ex}
\begin{align}
r(c) &= \sum\limits_{i=1}^N\left({\|x_i-c\|- R_{\text{mean}}(c)}\right)^{2}\,,\\
\hat{c} &= \argmin_{c \in \mathbb{R}^2}  r(c)\,.
\end{align}
An estimate of the quality of the fitted circle is given by the normalized residuals $r_{\text{norm}}(\hat{c}) = \sqrt{\frac{r(\hat{c})}{N}}$.
Detections can be filtered by a size threshold $\lambda_\text{radius}$ and a residual threshold parameter $\lambda_\text{res}$, which are empirically determined by evaluating recorded test data. Fitted circles $\hat{c}$ are accepted as valid detections if $R_{\text{mean}}(\hat{c}) < \lambda_\text{radius}$ and $r_{\text{norm}}(\hat{c}) < \lambda_\text{res}$.

The detector outputs 2D balloon centers and their radii in pixel coordinates.
A 3D position estimate can be calculated based on the balloon radius in the detections $R_{\text{mean}}$ and in real $R_{\text{real}} = \SI{30}{\centi\meter}$ which was fixed and known.
The additional information $R_{\text{real}}$ motivated us to estimate the depth of balloons without using the given depth image of our Intel RealSense Camera, since we observed noisy depth and a not negligible computation effort.
At first, we project the balloon center point $r_1 = (c_u, c_v, 1)$ and the point at the right balloon outline $r_2 = (c_u + R_{\text{mean}}, c_v, 1)$ into 3D camera coordinates at unit depth using the camera matrix \[
K=
\begin{bmatrix}
fx & 0 & cx \\
0 & fy & cy \\
0 & 0 & 1
\end{bmatrix}\,.
\]
The transformed points are $p_{1/2} = K^{-1} r_{1/2}$, with angle $\alpha = \arccos{(\frac{p_1 \cdot p_2}{\|p_1\| \|p_2\|})}$ between them. Further, the scalar $s$, which transforms $p_1$ from unit depth to the 3D balloon center point $P_\text{m} = s p_1$ in metric scaling is calculated as $s = \frac{R_{\text{real}}}{\tan{\alpha}}$.
The resulting 3D balloon center points are then further processed by an allocentric filter (\cf\refsec{sec:Balloon_Filter}).

The entire pipeline is very time efficient and runs with an average processing time of \SI{45}{\milli\second} per frame on the used Intel i7-6500U CPU with the Google Edge TPU connected over USB 3.0. Since the balloons are static, this computation time is more than sufficient as results during the contest confirmed. The quantization of the network for processing on the Edge TPU did not lead to any decrease in prediction quality. The small receptive field allows to easily detect multiple balloons in images as shown in~\reffig{fig:balloon_perception}.

Finetuning during the competition resulted in a significant decrease of background noise and enhanced the balloon outline detection of the network.
Consequently, parameters like residual threshold and minimum connected component size in postprocessing were adapted, so that balloons were detected even at large distances of up to \SI{50}{\meter}.

\bgroup
\begin{figure}[t]
	\centering
	\begin{subfigure}[b]{0.49\linewidth}
		\centering
		\includegraphics[width=\linewidth]{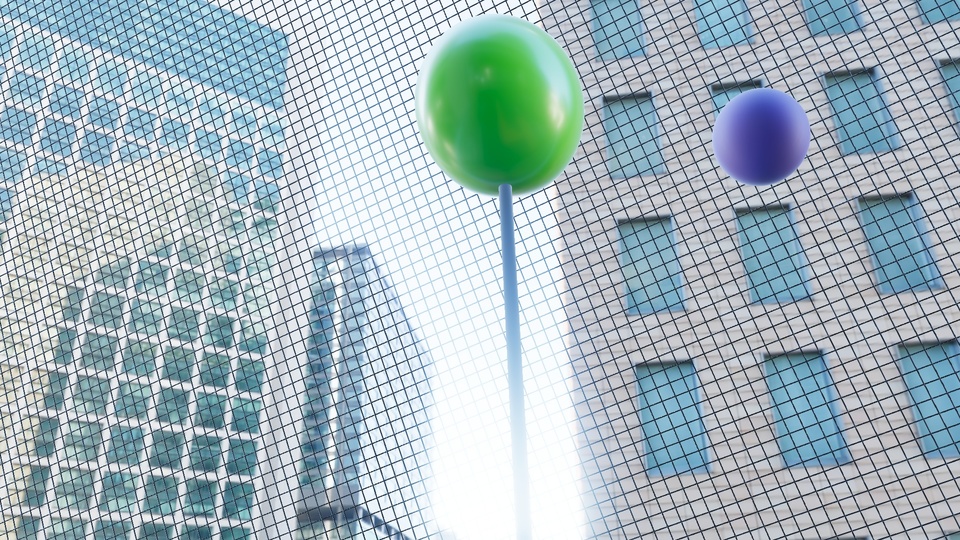}
		\label{fig:fig1}
	\end{subfigure}
	\begin{subfigure}[b]{0.49\linewidth}
		\centering
		\includegraphics[width=\linewidth]{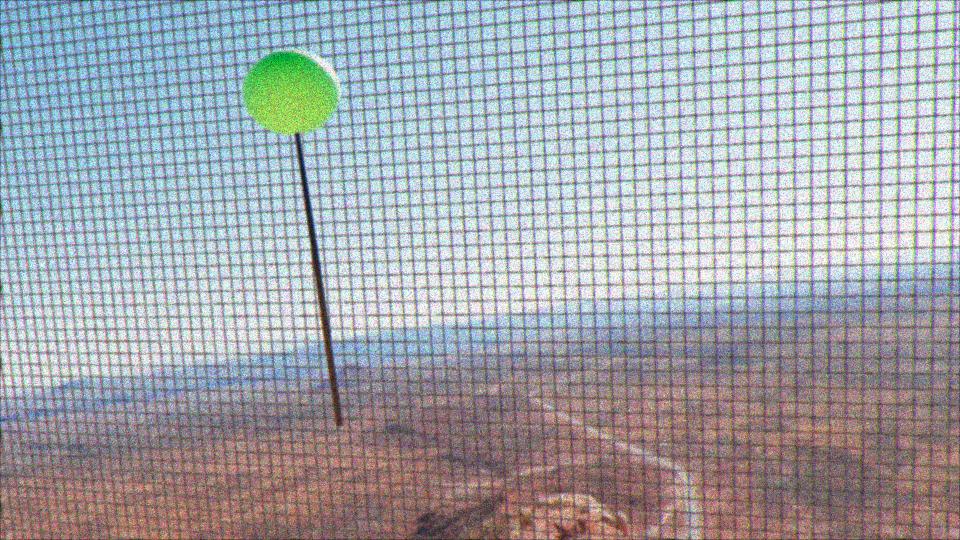}
		\label{fig:fig2}
	\end{subfigure}
    \vspace{-5.5ex}
	\caption{Synthetic images generated using EasyPBR~\cite{easy_pbr}.}
	\label{fig:ballon_synthetic}
\end{figure}
\egroup

\subsection{Balloon Filter}
\label{sec:Balloon_Filter}
For each image frame, the balloon perception (\refsec{sec:Balloon_Perception}) outputs a list of current balloon detections, described as egocentric 3D positions in camera coordinates.
These are processed by a filter to reject outliers and to aggregate them into a list of hypotheses $\mathcal H$ of possible balloon positions.
Each hypothesis $\mathcal H_i\in\mathcal H$ consists of
\begin{itemize}
 \item a history $\mathcal D_i := (d^i_1, \dots, d^i_8)$ of the last eight detections that were assigned to it,
 \item an estimate of the balloon position $P_i:=\frac{1}{|\mathcal D_i|}\sum_{d\in\mathcal D_i} d$, calculated as the running average over the detection history, and
 \item a counter for missed detections.
\end{itemize}
All hypotheses with at least eight detections are sorted with increasing distance to the current MAV position and forwarded to the state machine.
In the following, we describe the assignment of detections to the different hypotheses and the removal of hypotheses which were created by false detections based on the missed detection counter.

\begin{figure}[t]
  \centering
  \begin{tikzpicture}
    \node[inner sep=0] (image1) at ( 0.00, 2.50) {\includegraphics[trim=00mm 00mm 00mm 00mm,clip,width=0.49\linewidth]{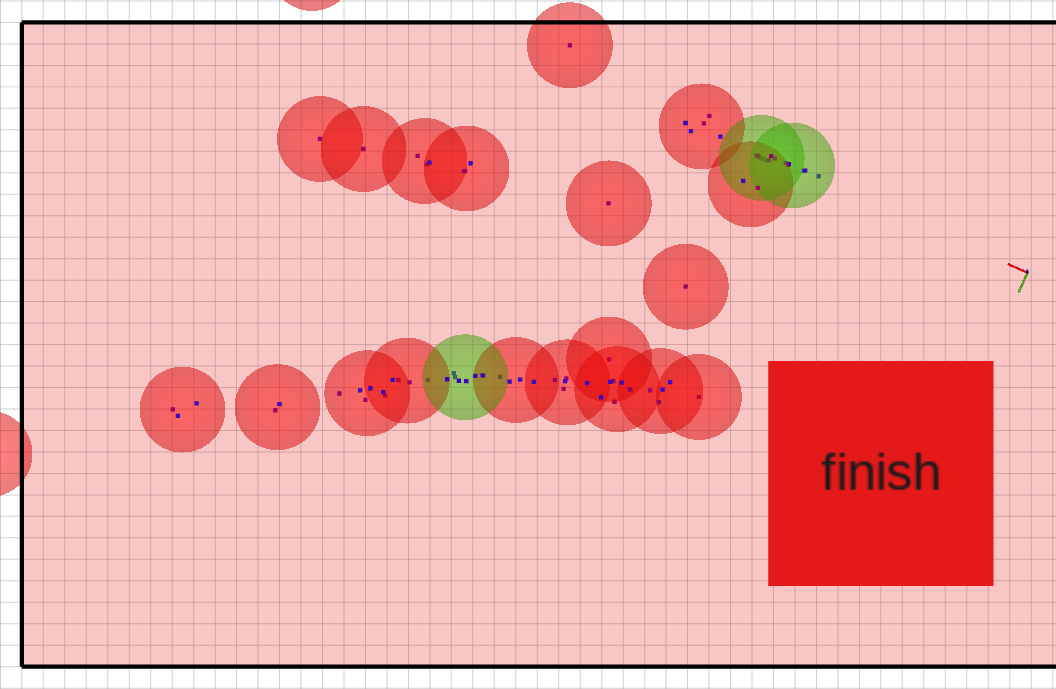}};
    \node[inner sep=0] (image2) at ( 4.40, 2.50) {\includegraphics[trim=00mm 00mm 00mm 00mm,clip,width=0.49\linewidth]{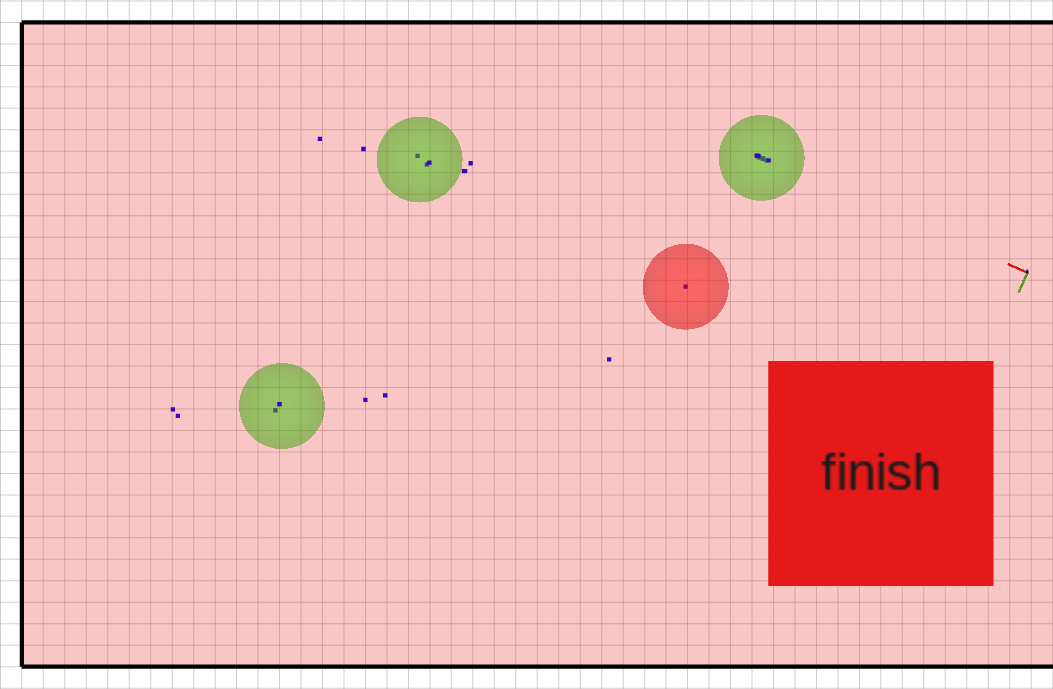}};
    \node[black,anchor=west] at ( 1.60, 3.00) {\tiny \textsf{MAV}};
    \node[black,anchor=west] at ( 6.00, 3.00) {\tiny \textsf{MAV}};
    \end{tikzpicture}
  \caption{Evaluation of different distance metrics. Detections are shown as blue dots. Hypotheses are shown as spheres, which are colored green if at least eight detections are assigned to them and red otherwise. \textbf{Left:} Euclidean distance on ground plane. \textbf{Right:} Distance to detection ray.}
  \label{fig:distance_metrics}
  \vspace{-2.5ex}
\end{figure}

\paragraph{Detection Assignment}
In a first step, the egocentric detections of the balloon detector are transformed into allocentric field coordinates.
Since the height of the balloons was predefined to be at \SI{2.5}{\meter}, all detections outside a height corridor from \SIrange[range-phrase = { to }]{1.5}{5.0}{\meter} are discarded.

For each remaining detection $d$, we determine the closest hypothesis $\mathcal H_{i^*}$ by minimizing the distance between the detection and all position estimates \ie choosing $i^*=\arg\min_i\{\textit{dist}(d,P_i)\}$.
If the distance is smaller than a threshold of \SI{2.0}{\meter}, we assign $d$ to $\mathcal H_{i^*}$, otherwise we create a new hypothesis.
Finally, hypotheses are merged when their estimated balloon positions become closer than \SI{2.0}{\meter}.

The choice of the distance measure $\textit{dist}(\cdot)$ is important to reduce the influence of detection noise and thus to achieve accurate assignments.
Since the center height of all balloons is fixed to $\SI{2.5}{\meter} + \frac{\SI{0.6}{\meter}}{2} = \SI{2.8}{\meter}$, $\textit{dist}(\cdot)$ can be chosen as the Euclidean distance on the ground plane to eliminate noisy height measurements.
However, due to the noisy depth estimation of the egocentric detections, this may result in multiple different hypotheses for the same balloon (Fig.~\ref{fig:distance_metrics} (left)).
Instead, we cast a ray $\tau$ in the direction of the detection and define $\textit{dist}(\cdot)$ to be the distance between the estimated balloon position and $\tau$.
This results in more accurate hypotheses assignments as shown in Fig.~\ref{fig:distance_metrics} (right).

\paragraph{Hypotheses Removal}
Once the MAV reaches a position above an estimated balloon, we assume the balloon to be popped and remove the corresponding hypothesis. If popping was not successful, the balloon will be detected again later and thus a new hypothesis for this balloon will be added (\cf \refsec{sec:Mission_Control_State_Machine}).

To further remove hypotheses created by false detections, we estimate the visibility of each hypothesis by projecting the estimated balloon coordinates onto the image plane of the camera.
Whenever a hypothesis is assumed to be visible but is not detected in the current frame, the corresponding missed detection counter is incremented.
If the number of missed detections exceeds a threshold of $30$, the corresponding hypothesis is removed.

\subsection{Laser Height Filter}
\label{sec:Laser_Height_Filter}

\begin{figure}[t]
  \centering
  \resizebox{1.0\linewidth}{!}{\begin{tikzpicture}[font=\sffamily,on grid,>={Stealth[inset=0pt,length=4pt,angle'=45]}]
\definecolor{red}{rgb}     {0.7,0.0,0.0}
\definecolor{green}{rgb}   {0.0,0.7,0.0}
\definecolor{blue}{rgb}    {0.0,0.0,0.7}
\definecolor{grey}{rgb}    {0.7,0.7,0.7}
\pgfplotsset{every tick label/.append style={font=\scriptsize}}
\begin{axis}[
  axis x line=center,
  axis y line=center,
  xticklabel={$\mathsf{\pgfmathprintnumber{\tick}}$},
  yticklabel={$\mathsf{\pgfmathprintnumber{\tick}}$},
  xtick={0,10,...,80},
  extra y ticks={1,2,...,6},
  extra tick style={grid=major},
  x label style={font=\scriptsize\sffamily, at={(axis description cs:0.5,-0.08)},anchor=north},
  y label style={font=\scriptsize\sffamily, at={(axis description cs:-0.08,.5)},rotate=90,anchor=south},
  y tick label style={/pgf/number format/.cd,fixed zerofill,precision=1,/tikz/.cd},
  xlabel={Time (s)},
  ylabel={Height (m)},
  height=5.5cm,
  xmin=0,
  ymin=0,
  ymax=6.5,
  legend style={font=\scriptsize\sffamily, at={(axis cs: 25,5.2)}, anchor=south west, row sep=-2.5pt, inner sep=0.8pt},
  legend image post style={scale=0.5},
  legend cell align={left},
  width=\linewidth]
 \addplot[color=blue,each nth point=2] 
	table[x=Time,y=Height] {data/run2/barometric.txt};
\addlegendentry{Barometer}
\addplot[color=red,each nth point=2] 
	table[x=Time,y=Height] {data/run2/laser.txt};
\addlegendentry{Laser~Height}
\addplot[color=black,each nth point=2] 
	table[x=Time,y=Height] {data/run2/output.txt};
\addlegendentry{Filter~Result}

\node[label,scale=1.0, align=center, rectangle, draw, fill=white, font=\scriptsize\sffamily] (drift) at (axis cs:9,5.5) {Barometer\\drift};
\draw[thick,->] (drift) -> (axis cs: 9,4.1);

\node[label,scale=1.0, align=center, rectangle, draw, fill=white, font=\scriptsize\sffamily] (invalid_laser) at (axis cs:48,1.2) {Invalid laser measurements};
\draw[thick,->] (invalid_laser) -> (axis cs: 28,1.2);

\node[label,scale=1.0, align=center, rectangle, draw, fill=white, font=\scriptsize\sffamily] (extrapolation) at (axis cs:62,6.1) {Extrapolated height};
\draw[thick,->] (extrapolation) -> (axis cs: 69,5.2);

\node[label,scale=1.0, align=center, rectangle, draw, fill=white, font=\scriptsize\sffamily] (interpolation) at (axis cs:60,2.7) {Linear convergence\\ below \SI{5}{\meter}};

\draw[thick,->] (interpolation) -> (axis cs: 71,4.8);
\end{axis}
\end{tikzpicture} }
  \vspace*{-4ex}
  \caption{Laser height correction. During takeoff, height estimation (black) is based on barometer data (blue) since laser measurements (red) are unreliable for too close distances. Once the MAV reaches an estimated height of \SI{1.0}{\meter}, height estimation is based on filtered laser data. Above \SI{5.0}{\meter} the laser becomes unreliable in the bright outdoor conditions and the height estimate is extrapolated using the change in barometric height measurements. Excerpt of Run~2.}
  \label{fig:height_eval}
\end{figure}
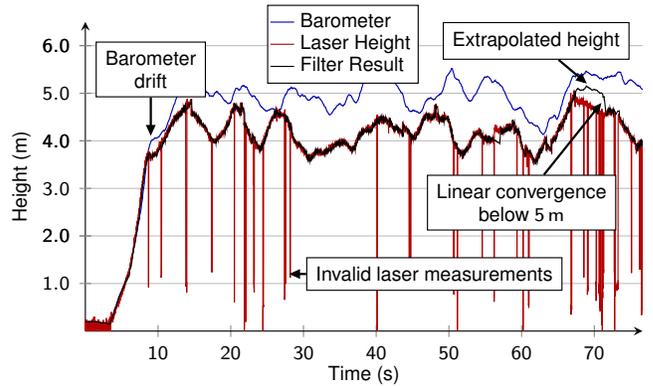

Precise height estimation can make the critical difference between popping a balloon, missing a balloon, or hitting a pole.
We therefore use measurements of a downwards facing LIDAR-Lite v3 as primary height source.

Detecting outliers and fusing the laser measurements with barometer data using an EKF would require to manually estimate sensor covariances.
However, we aim to use simple methods which allow fast debugging during the competition.
Thus, we choose an approach similar to the one that already proofed successful in MBZIRC 2017 (see \cite{ecmr2017_c3}).

A laser height filter determines whether the laser measurements are valid and thus can be used as height estimation.
However, when the laser measurements are assumed invalid, we extrapolate the latest height estimate using the change in the fused GNSS and barometric height.
As soon as the laser measurement is assumed valid again, we immediately correct our height estimate to the laser height or---if the extrapolated height drifted too much---linearly interpolate, allowing a maximum slope of \SI{1.5}{\meter\per\second}.

To determine whether a given laser measurement is valid, we check the following criteria:
\begin{itemize}
 \item Laser measurements below \SI{1}{\meter} are assumed to be invalid as they might be caused by measuring the spiked tentacles of our MAV.
 \item Laser measurements become unreliable for very low or large distances. Thus, we only consider laser measurements if the current height estimate is within \SIrange[range-phrase = { and }]{1.0}{5.0}{\meter}. Mind that this estimate might be extrapolated using GNSS/ barometric data.
 \item To reject outliers, we discard all laser measurements that differ more than \SI{15}{\centi\meter} from the latest valid laser measurement.
 The first valid laser measurement is bootstrapped by collecting ten measurements which pass the above criteria and selecting the one with the most inliers within a \SI{15}{\centi\meter} range. To be able to recover after a longer sequence of invalid measurements, this initialization process is repeated each time $100$ subsequent measurements (which corresponds to a time interval of \SI{1}{\second}) have been rejected.
\end{itemize}
\reffig{fig:height_eval} depicts height measurements and the filtered height estimates during Run~2.

\subsection{Trajectory Generation and Control}
\label{sec:Trajectory_Generation_and_Control}
Since the total time needed for the challenge is crucial and the MAV has to precisely hit the balloons, our method for trajectory generation and control is based on the method that already reliably worked during MBZIRC 2017 (see \cite{ecmr2017_c1} and \cite{ecmr2017_c3}).
The method is described in detail in~\cite{beul2016icuas} with the extensions from~\cite{beul2017icuas}. For reasons of brevity, in this section, we cover only the most important aspects of the algorithm.

\begin{table}[t]
\small
\caption{Parameters used at MBZIRC 2020.}
\label{tab:Parameters_used_at_MBZIRC_2020}
\vspace{-1.0em}
\begin{center}
\setlength{\tabcolsep}{1.9mm} 
\begin{tabular}{ccl|ccl}
  \toprule
  Parameter         & Axis & Value                                 & Parameter         & Axis              & Value                         \\
  \midrule
  $v_{max}$         & X,Y  & $\SI{5.0}{\meter\per\second}$         & $v_{max}$         & Z     & $\hphantom{0}\SI{1.0}{\meter\per\second}$ \\
  $a_{max}$         & X,Y  & $\SI{4.0}{\meter\per\second\squared}$ & $a_{max}$         & Z     & $\SI{10.0}{\meter\per\second\squared}$    \\
  $j_{max}$         & X,Y  & $\SI{5.0}{\meter\per\second\cubed}$   & $j_{max}$         & Z     & $\SI{50.0}{\meter\per\second\cubed}$      \\
  \bottomrule
\end{tabular}
\end{center}
\vspace{-2.5ex}
\end{table}

Based on a simple triple integrator model, our method analytically generates third-order time-optimal trajectories that satisfy input ($j_{min} \leq j \leq j_{max}$) and state constraints ($a_{min} \leq a \leq a_{max}$, $v_{min} \leq v \leq v_{max}$). Trajectories are computed from the current state $(p,v,a)_{MAV}^\intercal$ to the target state $(p,v,a)_{target}^\intercal$.
The X,Y, and Z axis are synchronized to arrive at the target state at the same time. By doing so, the MAV flies on a relatively straight path.

We directly use this trajectory generation method as a model predictive controller (MPC), running in a closed loop with $\SI{50}{\hertz}$. Our hardware does not support direct execution of sent jerk commands. We therefore assume pitch and roll to directly relate to $\theta = \atantwo(a_x,g)$ and $\phi = \atantwo(a_y,g)$. Thus, we send smooth pitch $\theta$ and roll $\phi$ commands for horizontal movement and smooth climb rates $v_z$ instead.
We use the parameters presented in \reftab{tab:Parameters_used_at_MBZIRC_2020}.

Although an arbitrary number of axes can be controlled by the above-men\-tioned method, we do not consider the yaw-axis $\Psi$ to be synchronized with the x, y and z-axis. For simplicity, we use proportional control for the yaw-axis. The yaw rate setpoint $\dot\Psi_{setp} = K_{p} \cdot (\Psi_{target} - \Psi_{MAV})$ with $K_{p} := 2.5$ for Run 1 and 2 and $K_{p} := 1.0$ for the Grand Challenge is sent to the MAV flight controller. The gain was reduced during the competition because of safety concerns (\cf~\refsec{sec:flight_path}).

\section{Evaluation}
\label{sec:Evaluation}
Our MAV system for balloon popping was operated in three competition runs during MBZIRC 2020, which are evaluated below. A video showcasing the evaluation of our Grand Challenge Run can be found on our website\footnote{\url{www.ais.uni-bonn.de/videos/ssrr_2020_mbzirc}}.

\subsection{Time Until Completion}
\label{sec:completion_time}

\begin{figure}[t]
	\centering
	\begin{tikzpicture}
        	\node[inner sep=0] (image1) at (0, 0){\includegraphics[width=\linewidth]{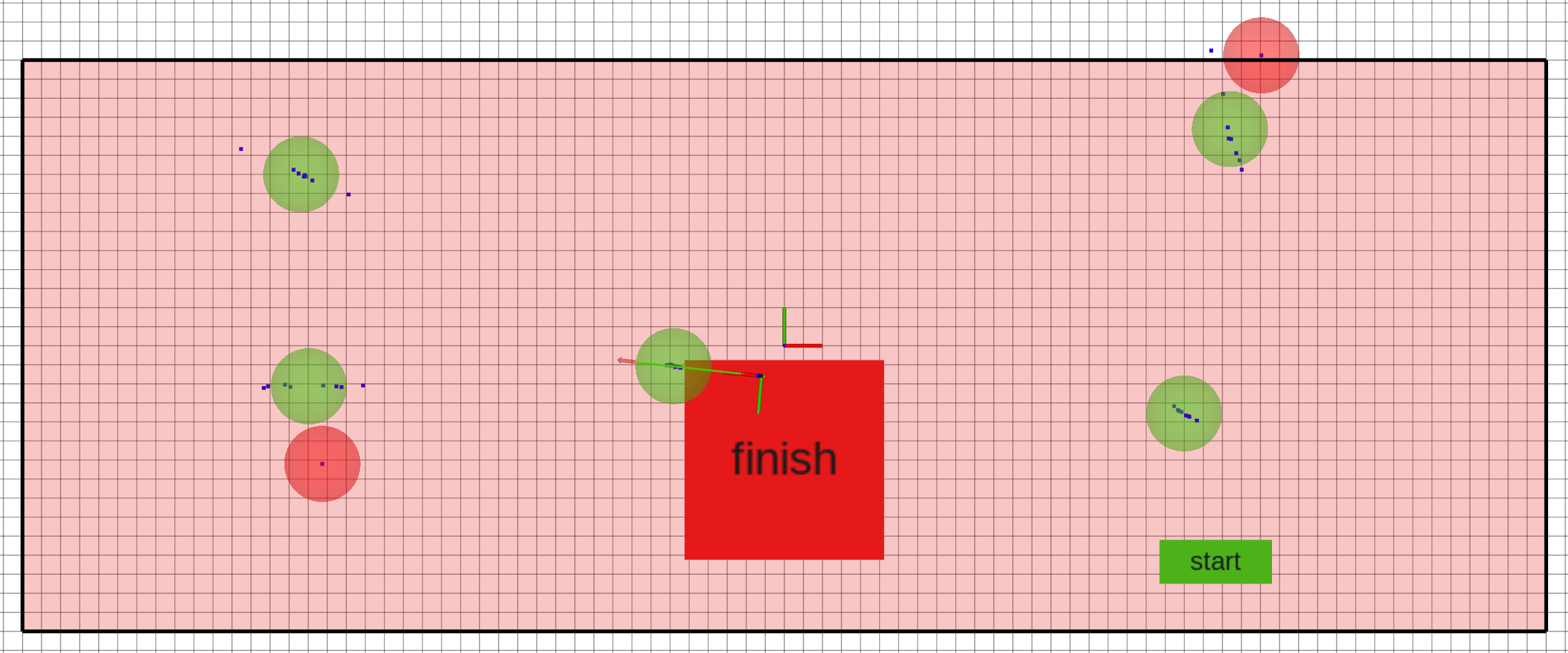}};
        	\node[black, anchor = south west] at ( 0.00,-0.15) {\tiny \textsf{field}};
        	\node[black] at ( 0.1,-0.35) {\tiny \textsf{MAV}};
    \end{tikzpicture}
	\caption{World model \SI{15}{\second} after takeoff during Run~2: Individual detections (blue squares) are accumulated into balloon hypotheses. All five balloons are detected (green circles). Outlier detections result in unconfirmed balloon hypotheses (red circles). The MAV is about to approach the first balloon. The goal pose (red arrow) is set right behind a balloon hypothesis. The field area is shaded in red, its origin is the \textit{field} frame.}
	\label{fig:world_model}
\vspace{-2.5ex}
\end{figure}

As our vision system detects the balloons at large distances (\cf~\refsec{sec:Balloon_Perception}), balloon hypotheses are added to the world model shortly after takeoff during all runs. During the second run, all five balloons were known to the filter only \SI{15}{\second} after takeoff (see~\reffig{fig:world_model}). In the other runs, only a fraction of the balloons was inside the field of view directly after takeoff. The known target hypotheses are approached right away, the MAV then flies on a search pattern for a short time only until it detects the remaining targets. The times at which the respective balloons were punctured are given in~\reftab{tab:timings}.\\
In the first run, two balloons were popped after \SI{30}{\second}. Then a reset occurred for \SI{8}{\minute}, as the MAV had gotten stuck in the Net at the arena borders due to an error in the GNSS-based geofencing. The challenge was completed after \SI{9}{\minute}~\SI{28}{\second}, but only \SI{1}{\minute}~\SI{28}{\second} flight time.\\
In the second run, the first two balloons were punctured right in sequence. Then, however, two balloons were missed---the puncturing did not work due to a suboptimal flight pattern (see~\refsec{sec:flight_path}). By repeating the search pattern and re-approaching the missed targets, in this run, all balloons were punctured after a total duration of \SI{1}{\minute}~\SI{40}{\second}.\\
In the final run during the Grand Challenge, all balloons were punctured in the first attempt. The time between two consecutive balloons were very similar, (\SIrange{12}{15}{\second}). Between Balloon~2 and 3, the MAV flew a search pattern to discover the remaining ones, which explains the longer time interval. The challenge was completed after a total time of \SI{1}{\minute}~\SI{21}{\second}, the shortest duration of all three competition runs.

\subsection{Flight Path}
\label{sec:flight_path}
\begin{figure}
    \centering
    \begin{subfigure}[t]{\linewidth}
        \centering
        \begin{tikzpicture}
            \node[inner sep=0] (image1) at (0, 0) {\includegraphics[width=\linewidth]{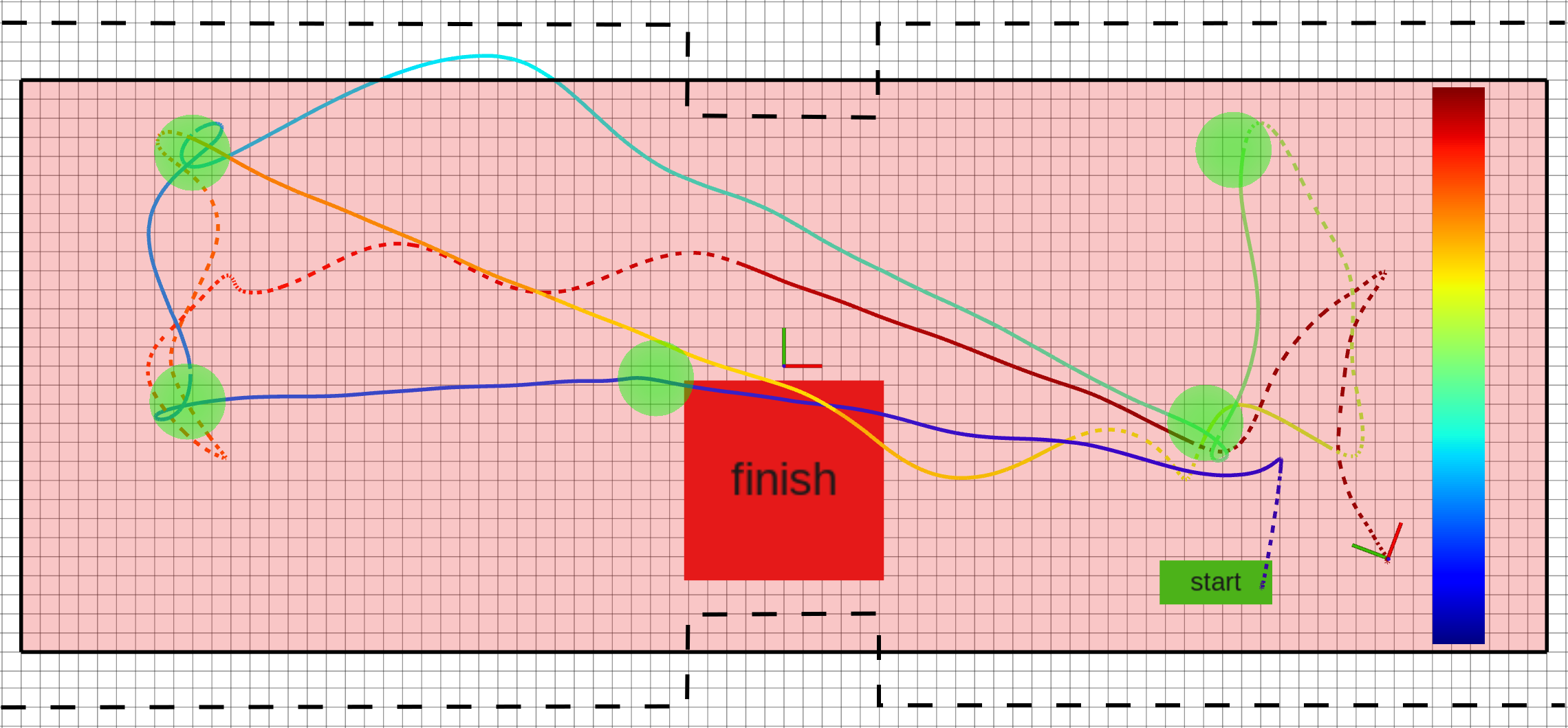}};
            \node[black] at (-0.70,-0.10) {\scriptsize\textbf{\textsf{1}}};
            \node[black] at (-3.30,-0.22) {\scriptsize\textbf{\textsf{2}}};
            \node[black] at ( 2.48, 1.18) {\scriptsize\textbf{\textsf{3}}};
            \node[black] at (-3.28, 1.15) {\scriptsize\textbf{\textsf{4}}};
            \node[black] at ( 2.32,-0.32) {\scriptsize\textbf{\textsf{5}}};
            \node[black,anchor=west] at ( 3.6,-1.45) {\tiny ---\textsf{0}};
            \node[black,anchor=west] at ( 3.6,-0.725) {\tiny ---\textsf{25}};
            \node[black,anchor=west] at ( 3.6, 0.00) {\tiny ---\textsf{50}};
            \node[black,anchor=west] at ( 3.6, 0.725) {\tiny ---\textsf{75}};
            \node[black,anchor=west] at ( 3.6, 1.45) {\tiny ---\textsf{100}};
            \node[black,anchor=south west] at (-0.05,-0.05) {\tiny \textsf{field}};
            \node[black,anchor=north] at ( 3.30,-1.0) {\tiny \textsf{MAV}};
        \end{tikzpicture}
        \caption{Run~2: The MAV chooses the direct path between consecutive balloons. In some cases, it turns directly above the balloons which prevents the piercing tentacles from working correctly. Balloons 4 and 5 need to be passed two resp. three times until successful puncturing. Flight path leaves allowed area at top left.}
    \end{subfigure}
    \par\medskip
    \begin{subfigure}[t]{\linewidth}
        \centering
        \begin{tikzpicture}
            \node[inner sep=0] (image1) at (0, 0) {\includegraphics[width=\linewidth]{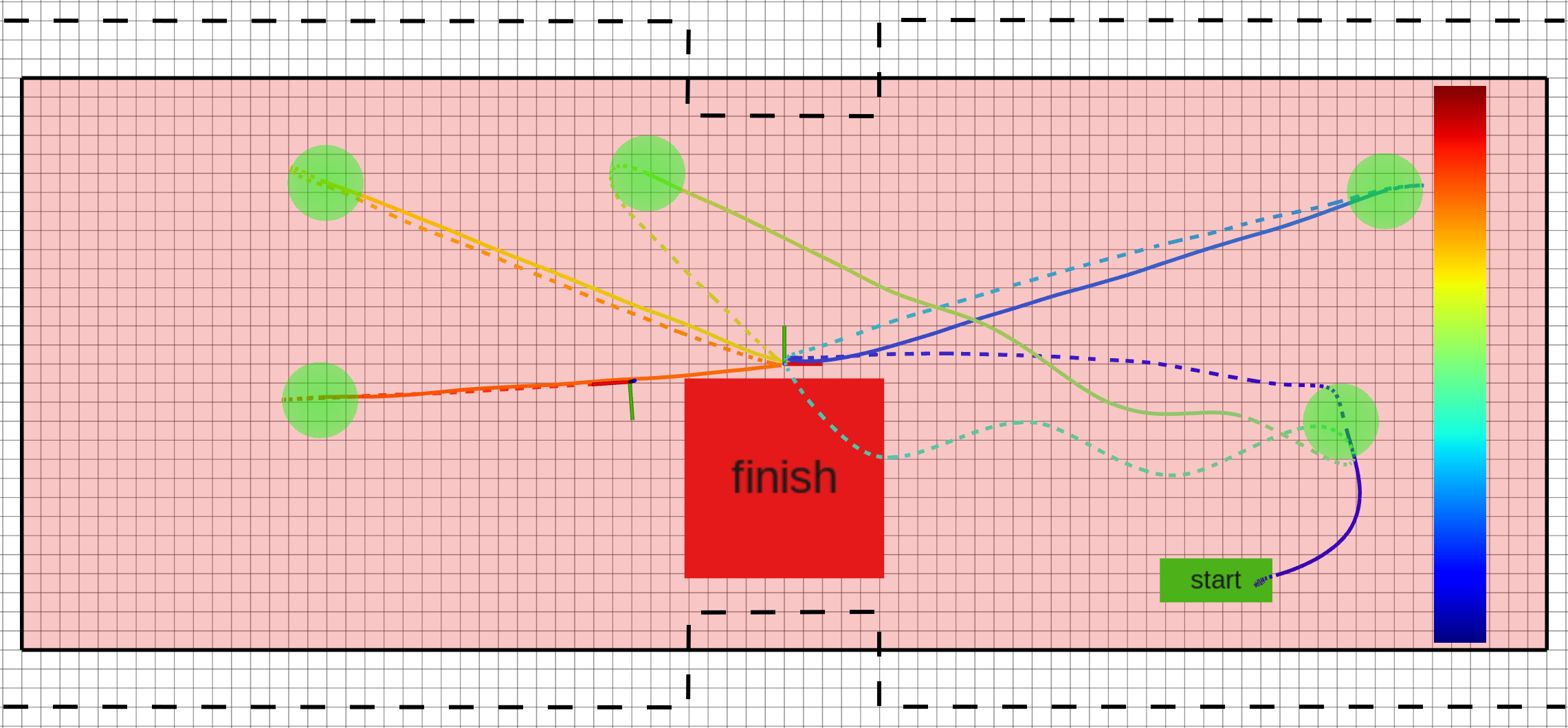}};
            \node[black] at ( 3.08,-0.35) {\scriptsize\textbf{\textsf{1}}};
            \node[black] at ( 3.30, 0.95) {\scriptsize\textbf{\textsf{2}}};
            \node[black] at (-0.75, 1.05) {\scriptsize\textbf{\textsf{3}}};
            \node[black] at (-2.52, 1.00) {\scriptsize\textbf{\textsf{4}}};
            \node[black] at (-2.55,-0.20) {\scriptsize\textbf{\textsf{5}}};
            \node[black,anchor=west] at ( 3.6,-1.45) {\tiny ---\textsf{0}};
            \node[black,anchor=west] at ( 3.6,-0.725) {\tiny ---\textsf{25}};
            \node[black,anchor=west] at ( 3.6, 0.00) {\tiny ---\textsf{50}};
            \node[black,anchor=west] at ( 3.6, 0.725) {\tiny ---\textsf{75}};
            \node[black,anchor=west] at ( 3.6, 1.45) {\tiny ---\textsf{100}};
            \node[black,anchor=south west] at (-0.05,-0.05) {\tiny \textsf{field}};
            \node[black,anchor=south] at (-0.9, -0.15) {\tiny \textsf{MAV}};
        \end{tikzpicture}
        \caption{Run~3: The MAV passes through the arena center after each balloon. It moves straight through the balloons, which leads to them being pierced reliably at the first attempt. Paths close to the boundaries are avoided by this strategy.}
    \end{subfigure}
    \caption{Comparison of flight paths between Runs~2 and 3 (colored by time). Solid line: Pop mode, dashed line: Search mode. The allowed flying area is shaded in red, the physical arena boundaries are marked with a dashed black line. Balloon hypotheses are displayed as green circles.}
    \label{fig:flight_path}
    \vspace{-2.5ex}
\end{figure}
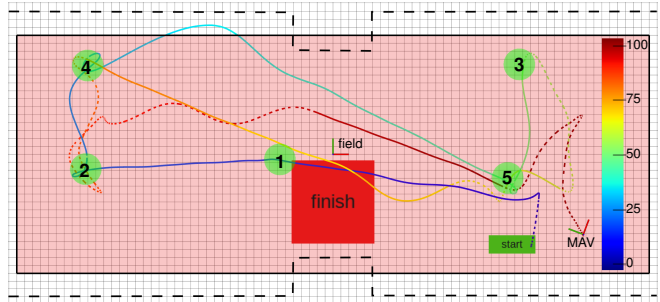
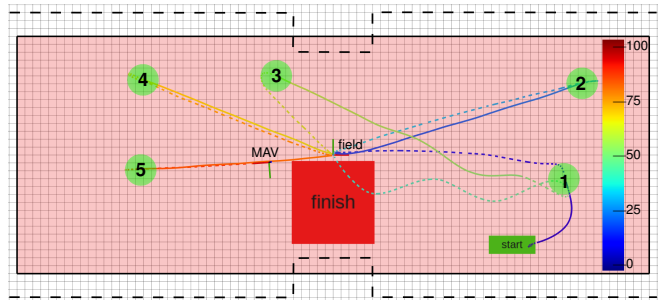

During the first and second run, the MAV always chose the direct path between two consecutive balloons. In the ideal case, this results in the shortest duration between two consecutive balloons (e.g. \SI{6}{\second} between the first and second balloon in Run~2). However, this can lead to the MAV flying dangerously close to the arena borders and could even have led to a crash during Run~2, had the controller chosen to pierce Balloon~3 and 4 in direct sequence. Our simple GNSS-based geofencing system which restricts the allowed flying area to a single rectangle could not correctly model the non-convex shape of the arena (see~\reffig{fig:flight_path}~(a)).
Furthermore, the above described behavior results in the MAV turning right over the balloons as it rotates to approach the next target. This is a suboptimal flight pattern as it prevents the dangling tentacles from piercing the balloons successfully. During the second run, two balloons were missed due to this suboptimal maneuvering and needed to be approached a second and even a third time.

\begin{table}[t]
\small
\caption{Timings of the balloon popping.}
\label{tab:timings}
\vspace{-1.0em}
\begin{center}
\setlength{\tabcolsep}{0.95mm} 
\begin{tabular}{l|ccccc}
  \toprule
                & Balloon~1        & Balloon~2        & Balloon~3                       & Balloon~4                                   & Balloon~5                        \\
  \midrule
  Run~1$^*$     & \SI{23}{\second} & \SI{30}{\second} & \SI{9}{\minute}~\SI{4}{\second} & \SI{9}{\minute}~\SI{11}{\second}            & \SI{9}{\minute}~\SI{28}{\second} \\
  Run~2         & \SI{17}{\second} & \SI{23}{\second} & \SI{51}{\second}                & \SI{1}{\minute}~\SI{15}{\second}            & \SI{1}{\minute}~\SI{40}{\second} \\
  Run~3         & \SI{11}{\second} & \SI{27}{\second} & \SI{56}{\second}                & \SI{1}{\minute}~\hphantom{0}\SI{8}{\second} & \SI{1}{\minute}~\SI{21}{\second} \\ 
  \bottomrule
\end{tabular}
\end{center}
\vspace{-1.0ex}
\quad($^*$) \SI{8}{\minute} reset time between 2nd and 3rd balloon.
\vspace{-2.5ex}
\end{table}

Turning above the balloons also had an impact on the safety of the flight maneuver. In Run~2, after popping Balloon~4, the MAV quickly turned while accelerating towards Balloon~5. Turning was so fast with $K_{p,yaw} = 2.5$, that the MAV was not able to redirect the acceleration fast enough due to the low jerk limit. Thus, the MAV first flew into a wrong direction---even entering the safety margin of the field---before correcting itself (see~\reffig{fig:flight_path}~(a) top-left). The trajectory rollout showed that the MAV was completely aware that it would leave the allowed area, but it simply could not compensate the erroneous acceleration fast enough. We prevented this possible safety hazard by setting $K_{p,yaw} = 1.0$, thus slowing down the yaw rate.

To overcome the remaining problems, an additional waypoint was added in the middle of the arena after each balloon for the final run. This results in a star-shaped flight pattern (see~\reffig{fig:flight_path}~(b)) and the balloons being passed in a straight line, without turning above them. Consequently, each balloon was pierced in the first attempt. The time between two consecutive balloons is slightly higher than it was before in the ideal case, but almost constant for each target, as no misses occur (see~\refsec{sec:completion_time}). Therefore, the challenge could be completed faster using the new behavior.
Moreover, the star-shaped flight pattern avoids trajectories close to the arena borders and leads to safe flight paths despite the non-convex arena outline without any additional obstacle avoidance system (\cf~\refsec{sec:Mission_Control_State_Machine}).

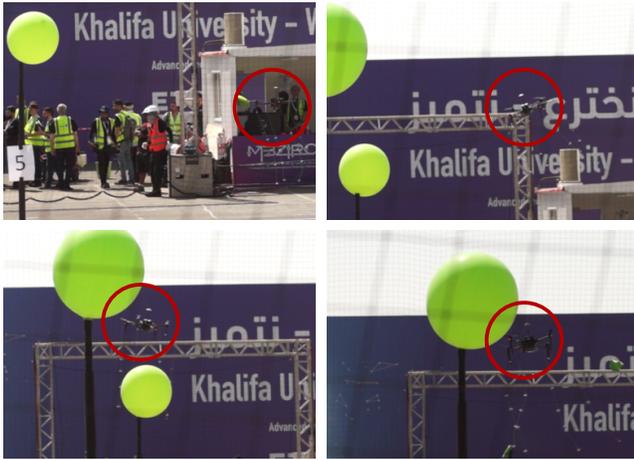
\begin{figure}[tb]
  \centering
  \begin{tikzpicture}
    \definecolor{red}{rgb}{0.7,0.0,0.0}   
    \node[inner sep=0] (image1) at (-2.15, 1.55) {\includegraphics[trim=3mm 28mm 125mm 30mm,clip,width=0.48\linewidth]{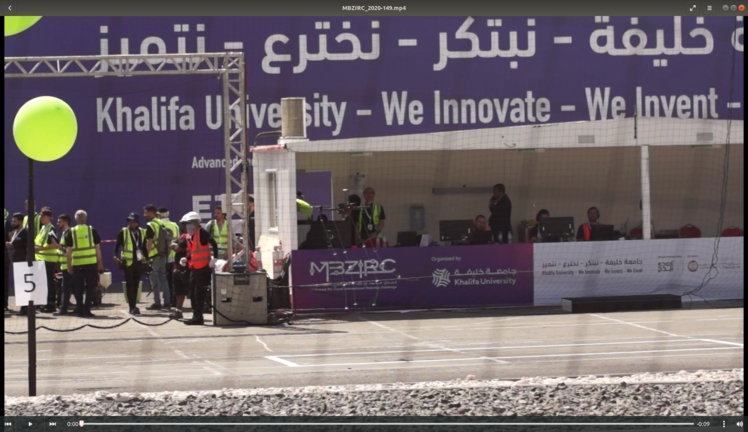}};
    \node[inner sep=0] (image2) at ( 2.15, 1.55) {\includegraphics[trim=3mm 28mm 125mm 30mm,clip,width=0.48\linewidth]{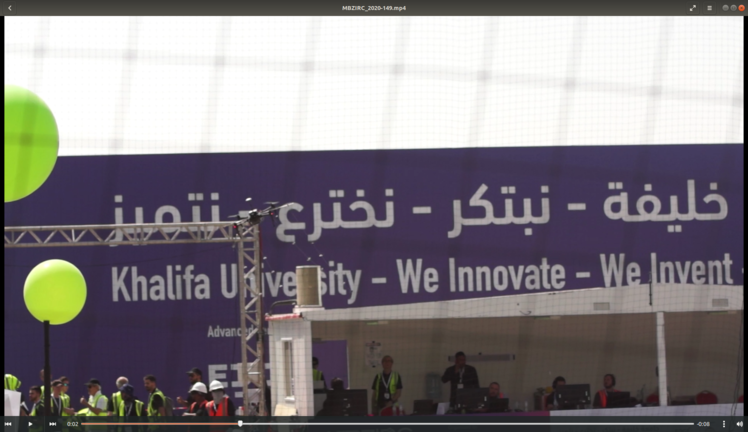}};
    \node[inner sep=0] (image3) at (-2.15,-1.55) {\includegraphics[trim=3mm 28mm 125mm 25mm,clip,width=0.48\linewidth]{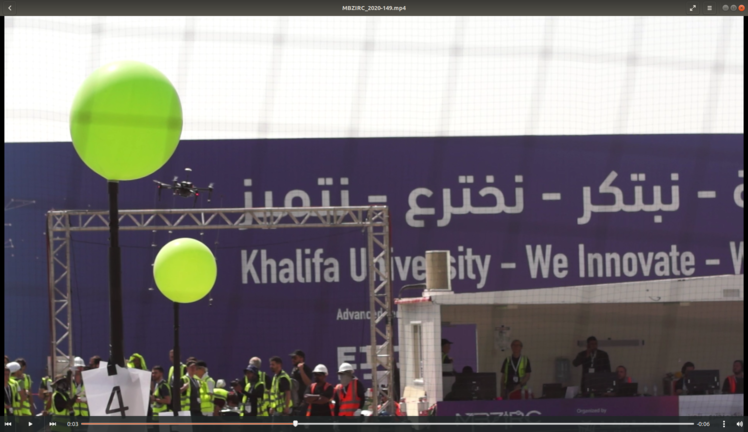}};
    \node[inner sep=0] (image4) at ( 2.15,-1.55) {\includegraphics[trim=3mm 28mm 125mm 25mm,clip,width=0.48\linewidth]{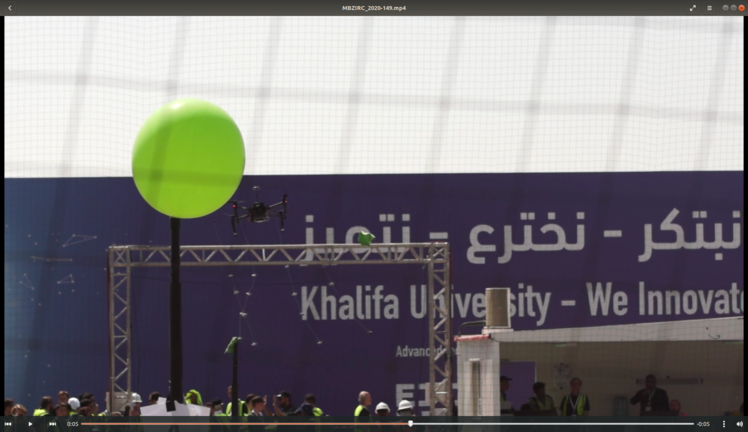}};
    \draw[line width=0.5mm, red] (-0.65, 1.60) circle (0.5cm and 0.5cm);
    \draw[line width=0.5mm, red] ( 2.70, 1.60) circle (0.5cm and 0.5cm);
    \draw[line width=0.5mm, red] (-2.40,-1.26) circle (0.5cm and 0.5cm);
    \draw[line width=0.5mm, red] ( 2.70,-1.50) circle (0.5cm and 0.5cm);
  \end{tikzpicture}
  \caption{Image sequence of popping the first balloon in the Grand Challenge (from top-left to bottom-right). The MAV (marked with the red circle) 1) takes off to \SI{4.0}{\meter}. 2) After detection of the first balloon, the MAV targets a position \SI{2.0}{\meter} behind and \SI{0.7}{\meter} above the center of the balloon. 3) It further accelerates to pass through the balloon with significant velocity. 4) It successfully pops the balloon. The entire shown process only takes \SI{4.9}{\second}.}
  \label{fig:popping_sequence}
\end{figure}

We show snapshots of the performance in~\reffig{fig:popping_sequence}. With this performance, we placed 5th in Challenge~1 including another subchallenge and 2nd in the Grand Challenge including five other subchallenges.

\section{Lessons learned}
As a retrospect towards our design choices, we deem the choice of using flexible tentacles that are dangling below the MAV to be a good decision. The flexible tentacles ensure that Jelly cannot get accidentally caught in the pole and additionally allow to have imperfect positioning with respect to the balloon. Competing teams that have experimented with rigid mechanism to pop the balloon struggled as the MAV would collide with the pole if the position estimate is not accurate enough. 

We also observed that multiple teams have opted to use a NVIDIA Jetson platform for running the MAV, including balloon detection networks~\cite{tony2020vision,Skyeye}. This might be an interesting direction to explore as the Edge TPUs, while powerful, are also limited in the architectures that can be executed on them.

\section{Conclusion}
\label{sec:Conclusion}

We have provided detailed insight into our robust MAV setup for quickly and robustly popping balloons. The viability of our approach has been demonstrated in a real-world scenario during the MBZIRC 2020 at which our MAV consistently performed as one of the fastest and most resilient among all competitors.

\par

In particular, the robust vision pipeline including the meticulous filtering of outliers combined with a robust fault-tolerant hardware and an overall simple system architecture that allowed for quick adjustments in the field made our approach to this challenge a huge success. The same counts for the laboriously handcrafted laser height filter that rejected a large number of outliers from the noisy sensor and a well-tested state machine that could be easily adapted during the individual runs. Also, our low-level control approach proofed to be reliable, precise and fast. It guided Jelly under real-world conditions, including noisy sensor information and external disturbances.

\par

We believe that our contribution, and in general all experience from the MBZIRC 2020, will inspire new ideas on how to operate flying robots in dynamic, real-world environments.

\bibliographystyle{IEEEtranBST/IEEEtran}
\bibliography{literature_references}

\begin{thebibliography}{10}
\providecommand{\url}[1]{#1}
\csname url@rmstyle\endcsname
\providecommand{\newblock}{\relax}
\providecommand{\bibinfo}[2]{#2}
\providecommand\BIBentrySTDinterwordspacing{\spaceskip=0pt\relax}
\providecommand\BIBentryALTinterwordstretchfactor{4}
\providecommand\BIBentryALTinterwordspacing{\spaceskip=\fontdimen2\font plus
\BIBentryALTinterwordstretchfactor\fontdimen3\font minus
  \fontdimen4\font\relax}
\providecommand\BIBforeignlanguage[2]{{%
\expandafter\ifx\csname l@#1\endcsname\relax
\typeout{** WARNING: IEEEtran.bst: No hyphenation pattern has been}%
\typeout{** loaded for the language `#1'. Using the pattern for}%
\typeout{** the default language instead.}%
\else
\language=\csname l@#1\endcsname
\fi
#2}}

\bibitem{MBZIRC2020}
MBZIRC, ``{MBZIRC} challenge description,''
  \url{https://www.mbzirc.com/challenge/2020}, 2019, accessed: 2020-03-16.

\bibitem{RoboCup_2019}
D.~Rodriguez, H.~Farazi, G.~Ficht, D.~Pavlichenko, A.~Brandenburger,
  M.~Hosseini, O.~Kosenko, M.~Schreiber, M.~Missura, and S.~Behnke, ``{RoboCup}
  2019 {AdultSize} winner {NimbRo}: Deep learning perception, in-walk kick,
  push recovery, and team play capabilities,'' \emph{RoboCup 2019: Robot World
  Cup XXIII}, pp. 631--645, 2019.

\bibitem{object_contour_2016}
J.~Yang, B.~Price, S.~Cohen, H.~Lee, and M.-H. Yang, ``Object contour detection
  with a fully convolutional encoder-decoder network,'' in \emph{{IEEE}
  {Conference} on {Computer} {Vision} and {Pattern} {Recognition} ({CVPR})},
  2016, pp. 193--202.

\bibitem{pascalvoc_2010}
M.~Everingham, L.~Van~Gool, C.~K.~I. Williams, J.~Winn, and A.~Zisserman, ``The
  {Pascal} {Visual} {Object} {Classes} ({VOC}) challenge,'' \emph{Int. Journal
  of Computer Vision}, vol.~88, no.~2, pp. 303--338, 2010.

\bibitem{conv_boundaries_2018}
K.-K. Maninis, J.~Pont-Tuset, P.~Arbel\'{a}ez, and L.~V. Gool, ``{Convolutional
  Oriented Boundaries: From Image Segmentation to High-Level Tasks},''
  \emph{{IEEE Transactions on Pattern Analysis and Machine Intelligence
  (TPAMI)}}, vol.~40, no.~4, pp. 819--833, 2018.

\bibitem{he_deep_2016}
K.~He, X.~Zhang, S.~Ren, and J.~Sun, ``Deep residual learning for image
  recognition,'' in \emph{IEEE Conference on Computer Vision and Pattern
  Recognition (CVPR)}, 2016.

\bibitem{mobilenet_2017}
A.~G. Howard, M.~Zhu, B.~Chen, D.~Kalenichenko, W.~Wang, T.~Weyand,
  M.~Andreetto, and H.~Adam, ``{MobileNets}: Efficient convolutional neural
  networks for mobile vision applications,'' 2017, arXiv:1704.04861.

\bibitem{mobilenetv2_2018}
M.~Sandler, A.~Howard, M.~Zhu, A.~Zhmoginov, and L.-C. Chen, ``{MobileNetV2}:
  Inverted residuals and linear bottlenecks,'' in \emph{{IEEE}/{CVF}
  {Conference} on {Computer} {Vision} and {Pattern} {Recognition}}, 2018.

\bibitem{edgetpu_usb}
Google, ``{EdgeTPU USB Accelerator},''
  \url{https://coral.ai/docs/accelerator/datasheet}, 2020, accessed:
  2020-03-20.

\bibitem{quantization_2018}
B.~Jacob, S.~Kligys, B.~Chen, M.~Zhu, M.~Tang, A.~Howard, H.~Adam, and
  D.~Kalenichenko, ``Quantization and training of neural networks for efficient
  integer-arithmetic-only inference,'' in \emph{{IEEE}/{CVF} {Conference} on
  {Computer} {Vision} and {Pattern} {Recognition} (CVPR)}, 2018.

\bibitem{Baca2017}
T.~Baca, P.~Stepan, and M.~Saska, ``Autonomous landing on a moving car with
  unmanned aerial vehicle,'' in \emph{European Conf. on Mobile Robots (ECMR)},
  2017.

\bibitem{Cantelli2017}
L.~Cantelli, D.~Guastella, C.~D. Melita, G.~Muscato, S.~Battiato, F.~D'Urso,
  G.~M. Farinella, A.~Ortis, and C.~Santoro, ``Autonomous landing of a {UAV} on
  a moving vehicle for the {MBZIRC},'' in \emph{20th International Conference
  on Climbing and Walking Robots and the Support Technologies for Mobile
  Machines (CLAWAR)}, 2017.

\bibitem{Battiato2017}
S.~Battiato, L.~Cantelli, F.~D'Urso, G.~M. Farinella, L.~Guarnera,
  D.~Guastella, C.~D. Melita, G.~Muscato, A.~Ortis, F.~Ragusa, and C.~Santoro,
  ``A system for autonomous landing of a {UAV} on a moving vehicle,'' in
  \emph{Image Analysis and Processing (ICIAP)}, 2017.

\bibitem{Falanga_SSRR2017}
D.~Falanga, A.~Zanchettin, A.~Simovic, J.~Delmerico, and D.~Scaramuzza,
  ``Vision-based autonomous quadrotor landing on a moving platform,'' in
  \emph{IEEE International Symposium on Safety, Security and Rescue Robotics
  (SSRR)}, 2017.

\bibitem{Ezair2014}
B.~Ezair, T.~Tassa, and Z.~Shiller, ``Planning high order trajectories with
  general initial and final conditions and asymmetric bounds,'' \emph{The Int.
  J. of Robotics Research}, vol.~33, no.~6, pp. 898--916, 2014.

\bibitem{realsense_d415}
Intel, ``{Depth Camera D415 --- Intel Realsense Depth and Tracking Cameras},''
  \url{https://www.intelrealsense.com/depth-camera-d415}, 2020, accessed:
  2020-03-20.

\bibitem{garmin_lidarlitev3}
Garmin, ``{Garmin Lidar-Lite v3},''
  \url{https://buy.garmin.com/de-DE/DE/p/557294}, 2020, accessed: 2020-03-20.

\bibitem{nimbro_networking}
M.~Schwarz, T.~Rodehutskors, D.~Droeschel, M.~Beul, M.~Schreiber, N.~Araslanov,
  I.~Ivanov, C.~Lenz, J.~Razlaw, S.~Sch\"uller, D.~Schwarz,
  A.~Topalidou-Kyniazopoulou, and S.~Behnke, ``{NimbRo Rescue}: Solving
  disaster-response tasks through mobile manipulation robot {Momaro},''
  \emph{J. of Field Robotics}, vol.~34, no.~2, pp. 400--425, 2017.

\bibitem{easy_pbr}
R.~A. Rosu, ``{EasyPBR},'' \url{https://github.com/RaduAlexandru/easy_pbr},
  2020, accessed: 2020-03-20.

\bibitem{Carlson:ECCV18}
A.~Carlson, K.~A. Skinner, R.~Vasudevan, and M.~Johnson{-}Roberson, ``Modeling
  camera effects to improve visual learning from synthetic data,'' in
  \emph{Computer Vision - {ECCV} Workshops}, 2018.

\bibitem{ecmr2017_c3}
M.~Nieuwenhuisen, M.~Beul, R.~A. Rosu, J.~Quenzel, D.~Pavlichenko, S.~Houben,
  and S.~Behnke, ``Collaborative object picking and delivery with a team of
  micro aerial vehicles at {MBZIRC},'' in \emph{European Conf. on Mobile Robots
  (ECMR)}, 2017.

\bibitem{ecmr2017_c1}
M.~Beul, S.~Houben, M.~Nieuwenhuisen, and S.~Behnke, ``Fast autonomous landing
  on a moving target at {MBZIRC},'' in \emph{European Conf. on Mobile Robots
  (ECMR)}, 2017.

\bibitem{beul2016icuas}
M.~Beul and S.~Behnke, ``Analytical time-optimal trajectory generation and
  control for multirotors,'' in \emph{Int. Conf. on Unmanned Aircraft Systems
  (ICUAS)}, 2016.

\bibitem{beul2017icuas}
------, ``Fast full state trajectory generation for multirotors,'' in
  \emph{Int. Conf. on Unmanned Aircraft Systems (ICUAS)}, 2017.

\bibitem{tony2020vision}
L.~A. Tony, S.~Jana, A.~Bhise, V.~BV, M.~S. Gadde, D.~Ghose, R.~Krishnapuram,
  \emph{et~al.}, ``Vision based target interception using aerial
  manipulation,'' \emph{arXiv:2009.13066}, 2020.

\bibitem{Skyeye}
R.~Suarez~Fernandez, A.~Rodríguez~Ramos, A.~Alvarez, J.~Rodríguez-Vazquez,
  H.~Bavle, L.~Lu, M.~Fernandez-Cortizas, A.~Rodelgo, A.~Cobano, D.~Alejo,
  D.~Acedo, R.~Rey, S.~Martinez-Rozas, M.~Molina, L.~Merino, F.~Caballero, and
  P.~Campoy, ``The {Skyeye} {Team} participation in the 2020 {Mohamed} {Bin}
  {Zayed} {International} {Robotics} {Challenge},'' 02 2020.

\end{thebibliography}

\end{document}